\documentclass[letterpaper,journal]{IEEEtran}
\usepackage{amsmath,amsfonts}
\usepackage{algorithmic}
\usepackage{algorithm}
\usepackage{array}
\usepackage{textcomp}
\usepackage{stfloats}
\usepackage{url}
\usepackage{verbatim}
\usepackage{graphicx}
\usepackage{cite}
\usepackage{hyperref}  
\hypersetup{
    colorlinks=true,      
    linkcolor=blue,       
    citecolor=blue,       
    urlcolor=blue,        
}
\usepackage{booktabs}
\usepackage{multirow}
\usepackage{tabularx}
\usepackage{pifont}
\usepackage{makecell}

\usepackage{amssymb}
\usepackage{mathrsfs}

\usepackage{subfigure}

\usepackage{bigstrut}
\usepackage{pifont}

\usepackage{orcidlink}

\usepackage{amsfonts}       
\usepackage{nicefrac}       
\usepackage{microtype}      
\usepackage{xcolor}         
\usepackage{paralist}
\usepackage{enumitem}
\usepackage{caption}
\usepackage[switch]{lineno}
\usepackage{siunitx}
\usepackage{newtxtext,newtxmath}

\usepackage[table]{xcolor} 

\begin{document}

\title{LC-TSRNet: A Land-Cover-Guided Two-Stage Network for Annual MODIS Surface Reflectance Time Series Reconstruction}

\author{Wenjuan~Zhang$^{\orcidlink{0000-0002-0534-0974}}$,~\IEEEmembership{Member,~IEEE},
Xuesong~Ma$^{\orcidlink{0009-0002-2329-3750}}$, 
Zhen~Li$^{\orcidlink{0000-0002-2049-2108}}$, 
Meixiang~Wang$^{\orcidlink{0009-0009-3620-3663}}$, 
Caimeng~Yi$^{\orcidlink{0009-0005-8677-7025}}$,
Jie~Pan$^{\orcidlink{0009-0001-2548-4684}}$,
Nan~Su$^{\orcidlink{0000-0001-9601-536X}}$,~\IEEEmembership{Member,~IEEE}, and
Bing~Zhang$^{\orcidlink{0000-0001-7311-9844}}$,~\IEEEmembership{Fellow,~IEEE}

\thanks{This work was supported by the Science and Disruptive Technology Program, AIRCAS under Grant 2024-AIRCAS-SDTP-11. (\textit{Corresponding author: Jie~Pan.})}

\thanks{Wenjuan Zhang, Jie Pan are with the State Key Laboratory of Remote Sensing and Digital Earth, Aerospace Information Research Institute, Chinese Academy of Sciences, Beijing 100101, China (e-mail: zhangwj@aircas.ac.cn; panjie@aircas.ac.cn).}

\thanks{Xuesong Ma, Caimeng Yi are with the Aerospace Information Research Institute, Chinese Academy of Sciences, Beijing 100094, China, and also with the College of Resources and Environment, University of Chinese Academy of Sciences, Beijing 100049, China (e-mail: maxuesong24@mails.ucas.ac.cn; yicaimeng25@mails.ucas.ac.cn).}

\thanks{Zhen Li is with the Aerospace Information Research Institute, Chinese Academy of Sciences, Beijing 100094, China (e-mail: lizhen02@aircas.ac.cn).}

\thanks{Meixiang Wang is with the School of Artificial Intelligence, Xidian University, Xi'an 710071, China (e-mail: 25171111438@stu.xidian.edu.cn).}

\thanks{Nan Su is with the College of Information and Communication Engineering, Harbin Engineering University, Harbin 150001, China, and also with the Key Laboratory of Advanced Marine Communication and Information Technology, Ministry of Industry and Information Technology, Harbin Engineering University, Harbin 150001, China (e-mail: sunan08@hrbeu.edu.cn).}

\thanks{Bing Zhang is with the State Key Laboratory of Remote Sensing and Digital Earth, Aerospace Information Research Institute, Chinese Academy of Sciences, Beijing 100101, China, and also with the College of Resources and Environment, University of Chinese Academy of Sciences, Beijing 100049, China (e-mail: zb@radi.ac.cn).}

}
\maketitle

\begin{abstract}

Moderate Resolution Imaging Spectroradiometer (MODIS) surface reflectance (SR) products often suffer from data gaps due to cloud contamination and sensor malfunctions, limiting their applications in long-term and large-scale studies. Most existing methods rely on cloud-free observations from optical time series or introduce synthetic aperture radar (SAR) time series as auxiliary information. However, due to the coarse spatial resolution of MODIS, cloud-free reference images are difficult to obtain, and SAR imagery is not well matched to the spatial resolution of MODIS, making it difficult to provide effective prior constraints. To address these issues, this study proposes a land-cover-guided two-stage time series reconstruction network (LC-TSRNet) for seamless reconstruction of annual MODIS SR products. In the first stage, an annual land cover map is introduced as prior knowledge, and a class-specific spectral prior module (CSPM) is designed by leveraging the spectral similarity within identical land cover types, thereby providing prior spectral constraints for missing regions. In the second stage, to address large-area, persistent, and overlapping gaps in MODIS imagery, a spatiotemporal multi-scale residual module (SMRM) is designed to adaptively exploit and integrate multi-scale spatiotemporal contextual information. Meanwhile, to address the limited collaborative utilization of spatio-temporal-spectral information, a three-dimensional attention module (3DAM) is designed to impose cross-dimensional joint constraints. Considering the lack of publicly available MODIS sample datasets, this study constructs the AIRMOD-TSR dataset, which contains annual MODIS SR time series and corresponding annual land cover maps. Extensive experimental results show that LC-TSRNet outperforms the comparison methods and achieves better reconstruction accuracy and spectral fidelity. Under the spatially adjacent scenario, LC-TSRNet improves PSNR by 0.56 dB and reduces SAM by 0.13$^\circ$. The vegetation monitoring application further verifies the practical value of the reconstructed SR time series. The dataset and code will be released at \url{https://github.com/WenjuanZhang-aircas/LC-TSRNet}.

\end{abstract}


\section{Introduction}
\label{sec:Intro}

\IEEEPARstart{O}{ptical} satellite remote sensing has emerged as a fundamental instrument for Earth observation. The periodic revisit of satellites enables the acquisition of time series imagery over the same region, supporting continuous monitoring and quantitative analysis of the Earth's surface \cite{fu2024remote}. Time series optical remote sensing data have been widely applied in vegetation monitoring \cite{brinkhoff2025empirical}, \cite{liu2025ndvi}, phenology extraction \cite{wu2021monitoring}, \cite{zhu2025improved}, land-use change analysis \cite{zhu2014continuous}, \cite{tollerud2023toward} and disaster response\cite{tulbure2022can}, \cite{zhao2025near}. Among various optical sensors, MODIS has become an important data source for long-term, large-scale time series studies because of its high temporal resolution, wide swath, and long-term continuous observations capabilities \cite{justice2002overview}. Among MODIS land products, the Surface Reflectance (SR) product serves as the most fundamental standardized data, providing critical inputs for higher-level products such as the Leaf Area Index (LAI), Fraction of Absorbed Photosynthetically Active Radiation (FPAR), and Net Primary Productivity (NPP). 

However, MODIS SR imagery commonly suffers from varying degrees of data gaps due to cloud contamination and sensor malfunctions \cite{shen2015missing}. Long-term continuous observations indicate that the average global cloud coverage for MODIS data reaches approximately 67\% \cite{king2013spatial}. This not only compromises the overall quality but also disrupts the temporal continuity of observations, thereby limiting the accuracy and reliability of applications. In addition, SR exhibits pronounced intra-annual dynamics driven by vegetation phenology and seasonal changes \cite{terasaki2025global}. Therefore, a time series reconstruction algorithm tailored to MODIS SR products is required to generate seamless annual SR time series while preserving intra-annual land surface dynamics.

To improve the data quality of optical remote sensing imagery, extensive research over the past decades has led to a variety of missing reconstruction methods. Early approaches can be categorized into spatial-based, temporal-based, and spatiotemporal hybrid methods \cite{shen2015missing}. However, most of these methods rely on linear assumptions \cite{zhang2025multi}. Consequently, they exhibit deficiencies in complex and fine-textured scenes, and are highly susceptible to the quality of auxiliary images.

In recent years, deep learning technology has been regarded as a new paradigm for missing reconstruction in remote sensing imagery. According to input configurations, deep-learning-based approaches can be categorized into single-image-based and sequence-based methods. Single-image-based methods typically reconstruct missing images by incorporating a nearby cloud-free optical image or a synthetic aperture radar (SAR) image as prior information \cite{ma2022deep}, \cite{wang2023cloud}, \cite{li2024hf}, \cite{pan2025m}. However, extending these methods to time series is challenging, as each missing image requires corresponding auxiliary data, and the correlations between sequential images are largely ignored.

Sequence-based methods typically take multi-temporal image sequences as input to fully exploit inter-image temporal dynamics and spatial correlations, thereby substantially improving reconstruction quality. Some studies use only cloud-contaminated optical sequences with missing observations, exploiting the cloud-free images to reconstruct the missing content \cite{zhang2020thick}, \cite{stucker2023u}, \cite{zhang2025multi}. Recently, TAMGAN \cite{wang2025temporal} uses a GAN framework with a temporal attention mechanism to reconstruct annual MODIS SR time series. Nevertheless, its global spectral noise filling strategy may struggle to handle completely missing observations, and its applicability to full 7-band MODIS SR products remains to be verified because only four bands are reconstructed. In addition, some studies address missing observations in optical time series imagery by integrating SAR data \cite{zhang2024u}, \cite{liu2025innovative}, \cite{shu2025114872}.

\begin{figure*}[htbp]
  \centering
  \includegraphics[width=\textwidth]{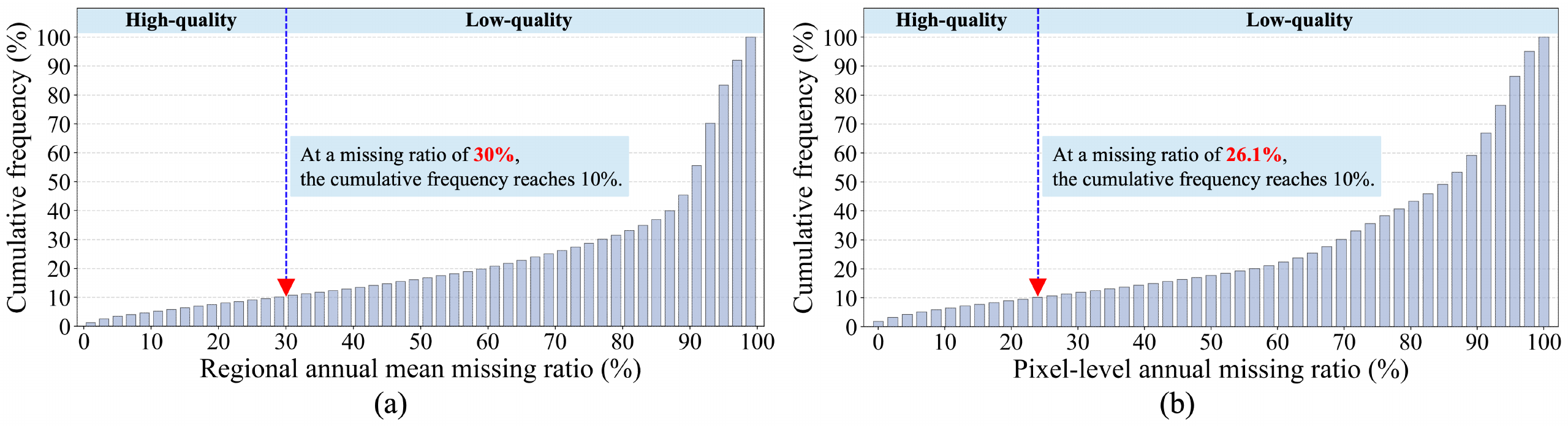}
  \caption{Cumulative frequency distributions of annual missing ratios in global MOD09A1 data during 2019--2023. (a) Regional annual mean missing ratio for 512 $\times$ 512-pixel regions. (b) Pixel-level annual missing ratio.}
  \label{fig:1GlobalMissing}
\end{figure*}

Although sequence-based deep learning methods have made substantial progress, MODIS SR time series reconstruction still faces the following challenges. First, most sequence-based methods rely on cloud-free observations from optical time series or introduce SAR time series as auxiliary data, and they have been developed and validated primarily on Sentinel and Landsat imagery \cite{zhang2025multi,tang2026anytimeformer}. However, due to the coarse spatial resolution of MODIS, such as 500 m for the MOD09A1 SR product, it is difficult to obtain cloud-free reference images for reconstruction at the regional scale \cite{krijger2007effect}. In this study, the global land surface was divided into 6854 regions of 512 $\times$ 512 pixels (approximately 65536~km$^2$), and the annual mean missing ratio of each region was derived from MOD09A1 data for 2019--2023. As shown in Fig.~\ref{fig:1GlobalMissing}(a), the resulting cumulative frequency distribution indicates that entirely cloud-free time series are absent at the regional scale, while regions with annual mean missing ratios below 10\% account for only 4.66\% of all global land regions. In addition, SAR imagery is generally acquired at spatial resolutions ranging from meters to tens of meters \cite{men2017high}, which are not naturally compatible with MODIS SR products. When SAR imagery is aggregated to the spatial resolution of MODIS, its fine-scale textural features and geometric structural details are substantially weakened \cite{schmitt2016data}, making it difficult to provide effective auxiliary information. Therefore, it is necessary to explore new prior constraints for MODIS SR time series reconstruction.

Second, current reconstruction methods typically employ fixed local receptive fields or static multi-scale structures for feature extraction, but lack an effective mechanism to adaptively exploit multi-scale spatiotemporal contextual information in response to varying extents of missing data \cite{liu2025glgf}, \cite{hou2025aca}. MODIS SR time series imagery often suffers from large-area gaps across multiple consecutive periods, under which valid information within local neighborhoods becomes severely insufficient \cite{suseno2025reconstructing}. As a result, these methods struggle to establish long-range spatiotemporal dependencies across missing regions, making reconstruction particularly challenging under large-area, persistent, and overlapping scenarios.

Third, existing methods still lack effective integration of spatial, temporal, and spectral information, thereby neglecting the intrinsic physical correlations among them \cite{zhang2026cloud}, \cite{zhang2026tssmamba}. MODIS SR time series imagery contains highly coupled spatial, temporal, and spectral information. Due to the lack of joint cross-dimensional constraint, these methods struggle to simultaneously preserve spatiotemporal continuity and spectral fidelity during reconstruction. 

Moreover, deep-learning-based reconstruction methods rely on extensive training samples. Existing publicly available satellite time series datasets for missing reconstruction are mainly developed for Sentinel and Landsat imagery, such as EarthNet2021 \cite{requena2021earthnet2021}, SEN12MS-CR-TS \cite{ebel2022sen12ms}, PASTIS-R \cite{garnot2022multi}, and AllClear \cite{zhou2024allclear}. In contrast, MODIS SR time series sample datasets remain scarce.

To address the above issues, this study proposes a novel \textbf{L}and-\textbf{C}over-guided two-stage \textbf{T}ime \textbf{S}eries \textbf{R}econstruction \textbf{Net}work (LC-TSRNet), which simultaneously reconstructs missing values across annual MOD09A1 8-day composite SR time series. The proposed LC-TSRNet adopts a coarse-to-fine two-stage architecture. In the first stage, the globally available MCD12Q1 IGBP annual land cover map is introduced as prior knowledge, and a class-specific spectral prior module (CSPM) is designed to impose spectral constraints, thereby generating a coarse representation. In the second stage, a spatiotemporal multi-scale residual module (SMRM) and a three-dimensional attention module (3DAM) are designed to refine spatio-temporal-spectral features. Finally, seamless SR time series with rich texture details are generated.

The main contributions of this study are as follows:

1) To impose prior constraints on MODIS SR time series reconstruction, we incorporate land cover information and design the CSPM. By leveraging the spectral similarity within identical land cover types at each time step, the CSPM transforms the semantic class knowledge into spectral prior knowledge, thereby imposing spectral-radiometric constraints on missing regions and improving the spectral fidelity of the reconstructed SR time series.

2) The SMRM is designed to adaptively exploit multi-scale spatiotemporal contextual information under varying extents of missing data. By employing parallel 3D convolutional branches with different dilation rates and a progressive aggregation mechanism, the SMRM enables cross-scale spatiotemporal interaction across missing regions, thereby providing effective joint spatiotemporal constraints for reconstructing large-area, persistent, and overlapping gaps.

3) The 3DAM is designed to adaptively enhance key features across the temporal, spatial, and spectral dimensions. By integrating three attention submodules in a parallel-cascaded structure, the 3DAM establishes joint cross-dimensional constraints and strengthens the intrinsic correlations among spatio-temporal-spectral features, thereby improving the spatiotemporal continuity and spectral fidelity of reconstructed SR time series.

4) To address the scarcity of publicly available MODIS SR time series datasets, the AIRMOD-TSR dataset is constructed. The dataset comprises annual MODIS SR time series and corresponding annual land cover maps for 13 representative regions of interest (ROIs) worldwide, providing a benchmark for MODIS SR time series reconstruction.

The rest of this article is organized as follows. Section~\ref{sec:Related} reviews related studies on missing reconstruction. Section~\ref{sec:Dataset} introduces the construction procedure of AIRMOD-TSR dataset. Section~\ref{Method} describes the proposed LC-TSRNet model and its key components in detail. Section~\ref{Experiments} presents the comparison results under different simulated scenarios and the ablation study. Section~\ref{Discussion} discusses the reconstruction performance in real-world scenarios, limitations, and future directions. Finally, Section~\ref{Conclusion} concludes this article.
\section{Related Works}
\label{sec:Related}

\subsection{Early Approaches}

Early research in missing reconstruction can generally be classified into three categories: spatial-based, temporal-based, and spatiotemporal hybrid methods.

\subsubsection{Spatial-Based Methods}

Spatial-based methods depend entirely upon the valid pixels within incomplete image, reconstructing gaps by exploiting image statistical characteristics and spatial correlations. Spatial interpolation assumes local continuity, inferring missing pixels by assigning weights to neighboring pixels \cite{tobler1970computer}. Typical methods include Kriging interpolation \cite{rossi1994kriging}, nearest-neighbor interpolation \cite{sibson1981brief}, and spline interpolation \cite{franke1982scattered}. Additionally, variation-based \cite{hardie1997joint, cheng2013inpainting}, exemplar-based \cite{criminisi2004region, he2014image}, and partial differential equation (PDE)-based methods \cite{bertalmio2000image} have also been applied to missing reconstruction. However, these methods are typically suitable only for small-area gaps or homogeneous regions. They exhibit significant limitations in accuracy and reliability when reconstructing large-area gaps or complex-texture regions.

\subsubsection{Temporal-Based Methods}

Temporal-based methods typically require auxiliary images from other dates and reconstruct gaps by exploiting temporal correlations across multi-temporal observations. Representative methods include temporal filtering \cite{beck2006improved}, histogram matching \cite{scaramuzza2005landsat}, and regression-based methods \cite{zeng2013recovering, li2014recovering}. For instance, Zeng et al. \cite{zeng2013recovering} proposed WLR, which combines sequential reference observations with a reference-free regularization framework. However, these methods typically require reference images that are temporally adjacent, cloud-free, and highly consistent in land-cover content.

\subsubsection{Spatiotemporal Hybrid Methods}

Spatiotemporal hybrid methods reconstruct missing area by combining spatial and temporal correlations across multi-temporal images. To address the missing data issue in Landsat ETM+ SLC-off imagery, Chen et al. \cite{chen2011simple} developed NSPI. Building on this foundation, Zhu et al. \cite{zhu2012modified} further proposed a modified NSPI (MNSPI) by refining similar-pixel selection and applying a weighted linear model for thick cloud removal. Integrating local temporal correlations with non-local spatial correlations, Li et al. \cite{li2016patch} proposed PM-MTGSR, a novel sparse representation approach driven by patch-matching. Ng et al.\cite{ng2017adaptive} proposed AWTC, which formulates gap filling as a weighted low-rank tensor recovery task. Chen et al. \cite{chen2017spatially} developed spatiotemporal weighted regression (STWR) to estimate missing pixels. However, such methods often struggle with fine-detail reconstruction in complex, textured scenes and typically involve a high computational cost.

\subsection{Deep-Learning-Based Approaches}

Deep-learning-based approaches can be broadly categorized into single-image-based and sequence-based methods according to their input configurations.

\subsubsection{Single-Image-Based Methods}

Single-image-based methods typically reconstruct missing observations by leveraging a nearby cloud-free optical image or a SAR image as prior information. For instance, Ma et al. \cite{ma2022deep} proposed TIIN, which leverages spatial priors from adjacent optical images to improve dehazing performance. 

In addition, SAR can penetrate clouds and haze, enabling surface observations under all-weather conditions, and is therefore widely used to guide optical image reconstruction. For instance, Wang et al. \cite{wang2023cloud} proposed USSRN-CR for thick cloud removal assisted by SAR data. Li et al. \cite{li2024hf} developed a two-stage optical-SAR fusion model (HF-T2CR) that simultaneously removes thin and thick clouds. Pan et al. \cite{pan2025m} introduced Mamba for thick cloud removal and proposed M\textsuperscript{3}-CR. Single image-based methods have achieved notable progress in missing reconstruction. However, these methods are not well suited for time series reconstruction, as they neglect sequential correlations and require corresponding auxiliary data for each missing image.

\subsubsection{Sequence-Based Methods}

\begin{figure*}[htbp]
  \centering
  \includegraphics[width=0.82\textwidth]{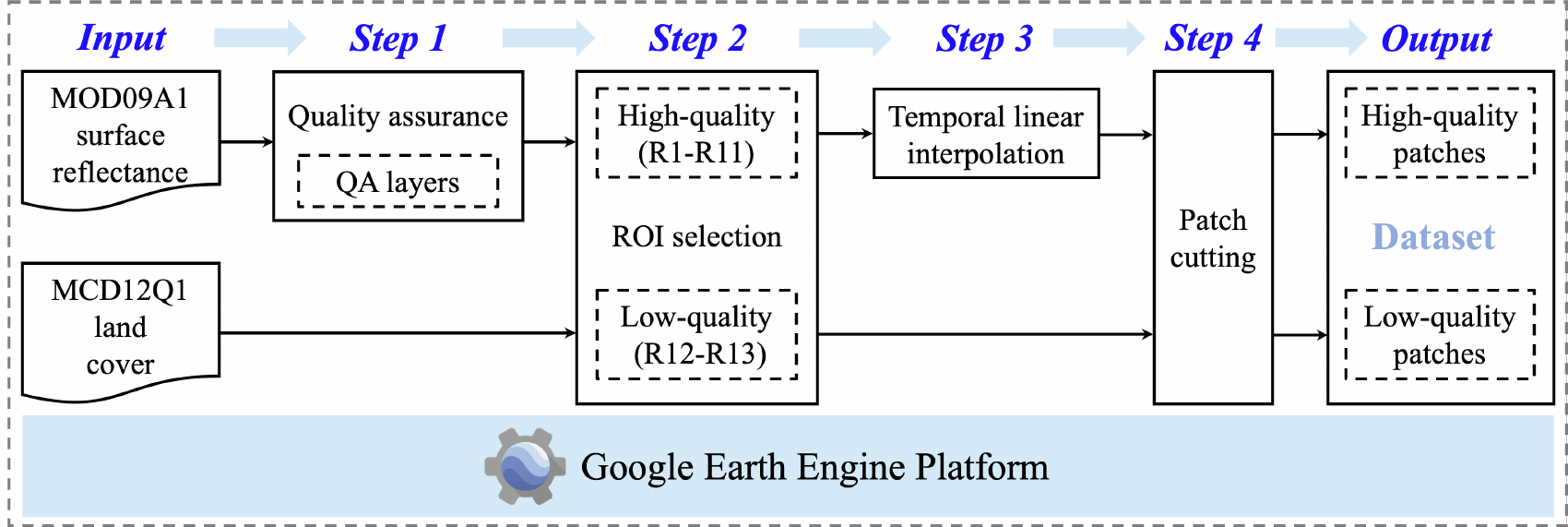}
  \caption{Overall workflow of data preprocessing.}
  \label{fig:2DataPre}
\end{figure*}

Sequence-based methods typically use multi-temporal image sequences as input to better exploit temporal dynamics and spatial correlations across observations. Some studies rely solely on optical image sequences with gaps to reconstruct the missing content. For example, Zhang et al.\cite{zhang2020thick} developed PSTCR, which constructs Sentinel-2 MSI and Landsat-8 OLI patch groups to extract spatiotemporal features and effectively remove thick cloud. Stucker et al. \cite{stucker2023u} proposed U-TILISE to reconstruct optical time series imagery. It extracts spatial features with 2D convolutions and temporal features with 1D self-attention, thereby reconstructing gaps in Sentinel-2 time series imagery. Zhang et al. \cite{zhang2025multi} proposed MS\textsuperscript{2}TAN, which leverages multi-temporal information to convert incomplete sequences into complete ones, and effectively validated it on Landsat and Sentinel-2 data. Wang et al.\cite{wang2025temporal} developed TAMGAN, a GAN-based framework that uses a 3D-convolutional generator for annual MODIS SR reconstruction.

Another category of methods addresses missing observations in optical time series imagery by incorporating SAR data. For instance, Zhang et al. \cite{zhang2024u} proposed U-SeqNet that combines U-Net with a sequence-to-sequence architecture and exploits SAR-optical multimodal information to restore Landsat-8 time series images. Liu et al. \cite{liu2025innovative} proposed STORI, which leverages a bidirectional recurrent architecture to integrate Sentinel-1 and Sentinel-2 time series for robust optical reconstruction in cloud-prone regions. Shu et al. \cite{shu2025114872} developed RESTORE-DiT, a conditional diffusion framework that jointly uses Sentinel-1 SAR and Sentinel-2 optical time series to reconstruct high-quality optical sequences. Tang et al. \cite{tang2026anytimeformer} developed AnytimeFormer, which fuses irregular Sentinel-2 optical observations with asynchronous Sentinel-1 SAR time series to reconstruct SR at arbitrary timestamps. However, methods based on optical sequences or optical-SAR sequences typically rely heavily on the quality of optical or SAR imagery.

\section{Materials and Dataset Construction}
\label{sec:Dataset}

In this study, the MOD09A1 8-day composite SR product was utilized, as it effectively captures seasonal dynamics while mitigating the severe data gaps inherent in daily observations \cite{xiao2005mapping}. For the MCD12Q1 annual land cover product, the IGBP classification scheme was adopted because of its widespread use in global studies \cite{yang2017accuracy}. Given the scarcity of publicly available MODIS SR time series datasets, AIRMOD-TSR was constructed using Google Earth Engine (GEE).

\subsection{Materials}

The MOD09A1 SR product is provided at 500~m spatial resolution, contains 46 composites per year, and includes seven bands spanning the visible to shortwave infrared (SWIR): B1 (620--670~nm), B2 (841--876~nm), B3 (459--479~nm), B4 (545--565~nm), B5 (1230--1250~nm), B6 (1628--1652~nm), and B7 (2105--2155~nm). Correspondingly, the MCD12Q1 product provides annual land cover maps at the same 500~m spatial resolution, which is spatially consistent with the MOD09A1 SR product and readily available, making it highly suitable as prior information for MODIS SR time series reconstruction.

\subsection{AIRMOD-TSR Construction}

As shown in Fig.~\ref{fig:1GlobalMissing}, the regional and pixel-level annual missing ratios of global MOD09A1 data during 2019--2023 were analyzed. The analysis indicates that severe data gaps are prevalent globally, rendering directly usable data extremely scarce. Consequently, we performed data preprocessing to construct AIRMOD-TSR for training and evaluation.

\subsubsection{Data Preprocessing}

\begin{figure*}[htbp]
  \centering
  \includegraphics[width=0.95\textwidth]{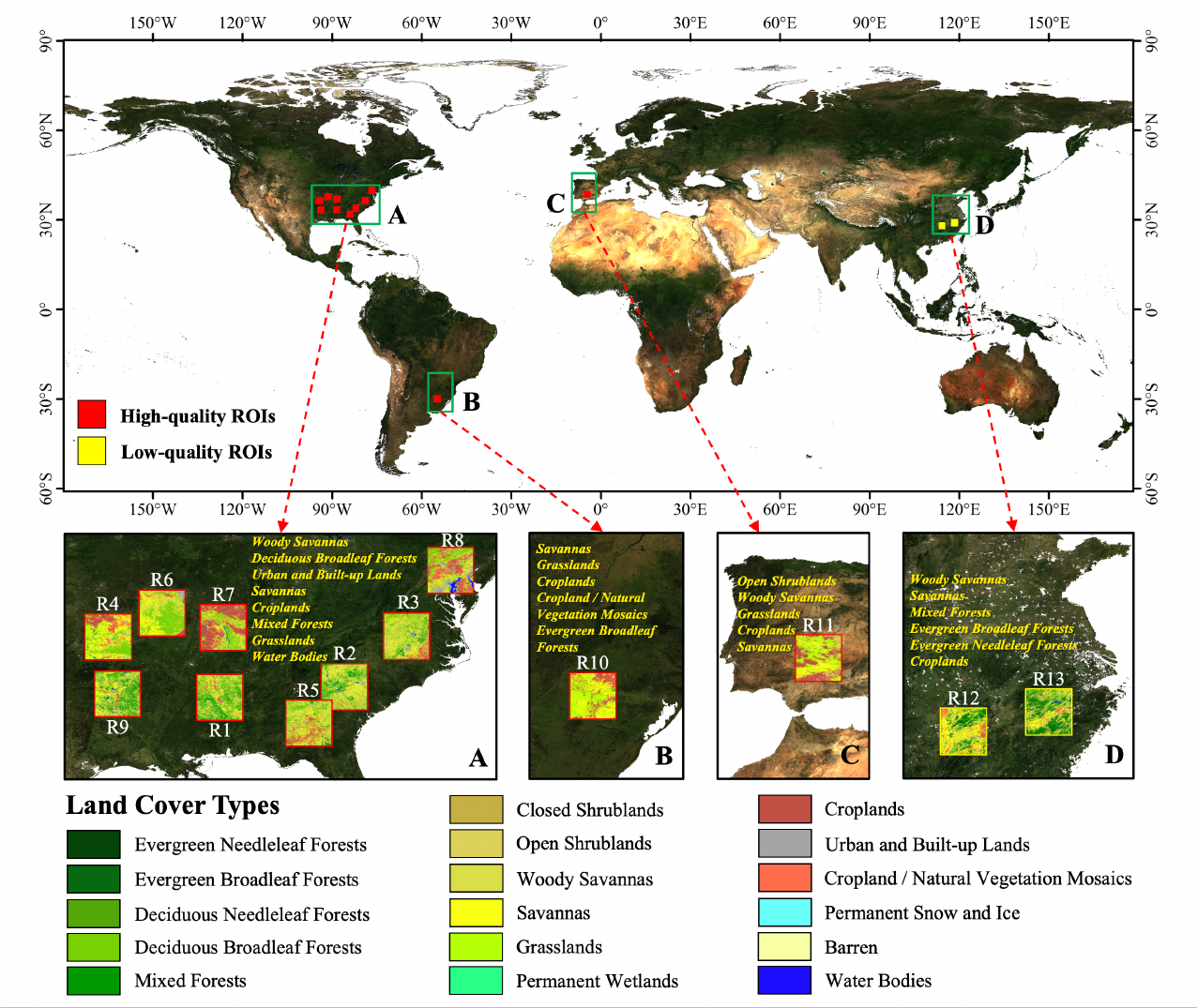}
  \caption{Geographic locations and land cover composition of the AIRMOD-TSR ROIs. The dataset consists of 13 globally distributed ROIs (R1--R13). Red and yellow boxes denote high- and low-quality ROIs, respectively.
  }
  \label{fig:3Dataset_Region}
\end{figure*}

As shown in Fig.~\ref{fig:2DataPre}, the overall workflow of data preprocessing includes four steps: quality assurance, ROI selection, temporal linear interpolation, and patch cutting.

\textbf{Step 1:} To ensure data reliability, composite-level quality filtering was first performed using the MOD09A1 quality assurance (QA) layers, with pixels affected by clouds, cloud shadows, aerosols, and snow/ice being masked.

\textbf{Step 2:} As shown in Fig.~\ref{fig:1GlobalMissing}, cumulative frequencies above 10\% are first reached at annual missing ratios of 30\% for regions and 26.1\% for pixels. Consequently, these values were adopted as thresholds for distinguishing high- and low-quality data. Based on this dual-level criterion, 11 high-quality ROIs and 2 low-quality ROIs were ultimately selected, each with a spatial size of 512 $\times$ 512 pixels.

\textbf{Step 3:} Although stringent quality control yields relatively reliable SR time series, small gaps inevitably remain even in high-quality data. In these ROIs, the mean longest consecutive gap per year is 2.81 composites. To construct spatiotemporally continuous ground-truth labels for model training, the remaining gaps were filled using temporal linear interpolation \cite{wang2025temporal}, formulated as:
\begin{equation}
    R_0 = R_b + \frac{T_0 - T_b}{T_a - T_b} (R_a - R_b)
    \label{eq:1linear_interp}
\end{equation}

\noindent where ${R_0}$ denotes the reflectance of the current pixel obtained by temporal linear interpolation; ${R_b}$ and ${R_a}$ denote the reflectance values at the nearest valid observation times before and after the current pixel in the time series, respectively; ${T_0}$, ${T_b}$, and ${T_a}$ denote the acquisition dates of ${R_0}$, ${R_b}$, and ${R_a}$, respectively, represented as the day of year (DOY).

\textbf{Step 4:} Finally, to balance training efficiency and computational resource constraints, a patch-based strategy was adopted, in which each sample image was divided into $64 \times 64$ patches with a 32-pixel overlap between adjacent patches.

\subsubsection{Description of AIRMOD-TSR}

AIRMOD-TSR comprises SR time series and corresponding annual land cover maps for 13 representative ROIs worldwide. The dataset contains 27 data groups in total, each corresponding to a full-year SR sequence for a single ROI. R1--R11 are high-quality ROIs, whereas R12--R13 are low-quality ROIs with corresponding missing masks provided. Each ROI covers $512 \times 512$ pixels, corresponding to approximately 65536~km$^2$, and contains more than 10 IGBP land cover classes. AIRMOD-TSR provides data in the World Geodetic System 1984 (WGS84) coordinate system at two spatial sizes, including complete $512 \times 512$ ROI data and $64 \times 64$ patches, with SR values scaled to [0, 10000]. Fig.~\ref{fig:3Dataset_Region} shows the geographical locations and land cover composition of these ROIs, while detailed information is provided in Table~\ref{tab:1DatasetRegion}.

\begin{table}[!t]
  \caption{Detailed information on the AIRMOD-TSR ROIs.}
  \label{tab:1DatasetRegion}
  \centering
  \footnotesize
  \renewcommand{\arraystretch}{1.2}
  \setlength{\tabcolsep}{7.5pt} 
  
  \begin{tabular}{lllll} 
    \toprule 
    
    \makecell[l]{Region \\ ID} & 
    \makecell[l]{Longitude \\ range} & 
    \makecell[l]{Latitude \\ range} & 
    \makecell[l]{Selected \\ year(s)} & 
    \makecell[l]{Annual mean \\ missing ratio} \\ 
    
    \midrule 

    \addlinespace[3pt] 
    \multicolumn{5}{c}{\textbf{High-quality ROIs}} \\ 
    \addlinespace[5pt] 

    R1  & \makecell[l]{87.31$^{\circ}$W--\\89.61$^{\circ}$W} & \makecell[l]{32.17$^{\circ}$N--\\34.47$^{\circ}$N} & \makecell[l]{2019, 2022, \\ 2023} & 20.78\% \\
    \addlinespace[3pt]
   
    R2  & \makecell[l]{80.83$^{\circ}$W--\\83.13$^{\circ}$W} & \makecell[l]{32.70$^{\circ}$N--\\35.00$^{\circ}$N} & \makecell[l]{2019, 2022, \\ 2023} & 20.26\% \\
    \addlinespace[3pt] 
    
    R3  & \makecell[l]{77.64$^{\circ}$W--\\79.94$^{\circ}$W} & \makecell[l]{35.32$^{\circ}$N--\\37.62$^{\circ}$N} & \makecell[l]{2019, 2020, \\ 2022, 2023} & 18.70\% \\
    \addlinespace[3pt]
    
    R4  & \makecell[l]{93.07$^{\circ}$W--\\95.37$^{\circ}$W} & \makecell[l]{35.29$^{\circ}$N--\\37.59$^{\circ}$N} & \makecell[l]{2019, 2022, \\ 2023} & 16.93\% \\
    \addlinespace[3pt]
    
    R5  & \makecell[l]{82.69$^{\circ}$W--\\84.99$^{\circ}$W} & \makecell[l]{30.83$^{\circ}$N--\\33.13$^{\circ}$N} & 2019, 2022 & 22.85\% \\
    \addlinespace[3pt]
    
    R6  & \makecell[l]{90.27$^{\circ}$W--\\92.57$^{\circ}$W} & \makecell[l]{36.53$^{\circ}$N--\\38.83$^{\circ}$N} & \makecell[l]{2019, 2022, \\ 2023} & 19.76\% \\
    \addlinespace[3pt]
    
    R7  & \makecell[l]{87.12$^{\circ}$W--\\89.42$^{\circ}$W} & \makecell[l]{35.80$^{\circ}$N--\\38.10$^{\circ}$N} & \makecell[l]{2019, 2022, \\ 2023} & 21.98\% \\
    \addlinespace[3pt]
    
    R8  & \makecell[l]{75.35$^{\circ}$W--\\77.65$^{\circ}$W} & \makecell[l]{38.77$^{\circ}$N--\\41.07$^{\circ}$N} & 2022 & 21.93\% \\
    
    \addlinespace[3pt] 
    
    R9  & \makecell[l]{92.59$^{\circ}$W--\\94.89$^{\circ}$W} & \makecell[l]{32.33$^{\circ}$N--\\34.63$^{\circ}$N} & 2022 & 15.00\% \\
    \addlinespace[3pt]
    
    R10 & \makecell[l]{53.67$^{\circ}$W--\\55.97$^{\circ}$W} & \makecell[l]{28.78$^{\circ}$S--\\31.08$^{\circ}$S} & 2022 & 11.85\% \\
    \addlinespace[3pt]
    
    R11 & \makecell[l]{3.49$^{\circ}$W--\\5.79$^{\circ}$W}  & \makecell[l]{37.54$^{\circ}$N--\\39.84$^{\circ}$N} & 2023 & 16.74\% \\
    \addlinespace[3pt]

    \multicolumn{5}{c}{\textbf{Low-quality ROIs}} \\
    \addlinespace[5pt]
    
    R12 & \makecell[l]{113.02$^{\circ}$E--\\115.32$^{\circ}$E} & \makecell[l]{26.93$^{\circ}$N--\\29.23$^{\circ}$N} & 2022 & 60.11\% \\
    
    \addlinespace[3pt]
    
    R13 & \makecell[l]{117.43$^{\circ}$E--\\119.73$^{\circ}$E} & \makecell[l]{27.88$^{\circ}$N--\\30.18$^{\circ}$N} & 2023 & 54.22\% \\
    
    \bottomrule
  \end{tabular}
\end{table}

R1--R9 are primarily located in the southeastern United States. These ROIs are characterized by a humid subtropical climate and complex landscape heterogeneity. The land cover composition is dominated by woody savannas, savannas, deciduous broadleaf forests, mixed forests, and croplands, interspersed with grasslands, urban areas, and water bodies.

R10 is located in southeastern South America (southern Brazil and northern Uruguay) and represents a typical Pampas landscape. It is dominated by savannas, grasslands, and croplands, with spatial patterns distinct from the training regions. As R10 lies in the opposite hemisphere from the other ROIs, it exhibits seasonal phase inversion, thereby increasing the difficulty of capturing phenological dynamics.

R11 is located in southern Spain (Andalusia and Castilla-La Mancha) and represents a typical Mediterranean landscape. Its land cover is dominated by grasslands, croplands, savannas, woody savannas, and open shrublands.

R12 and R13 are located in southern China under a subtropical humid monsoon climate, where persistent clouds and frequent precipitation cause severe missing observations in the MOD09A1 product. They feature highly heterogeneous and fragmented landscapes dominated by savannas, forests, and croplands.

\section{Methodology}
\label{Method}

\subsection{Overview}

\begin{figure*}[htbp]
  \centering
  \includegraphics[width=\textwidth]{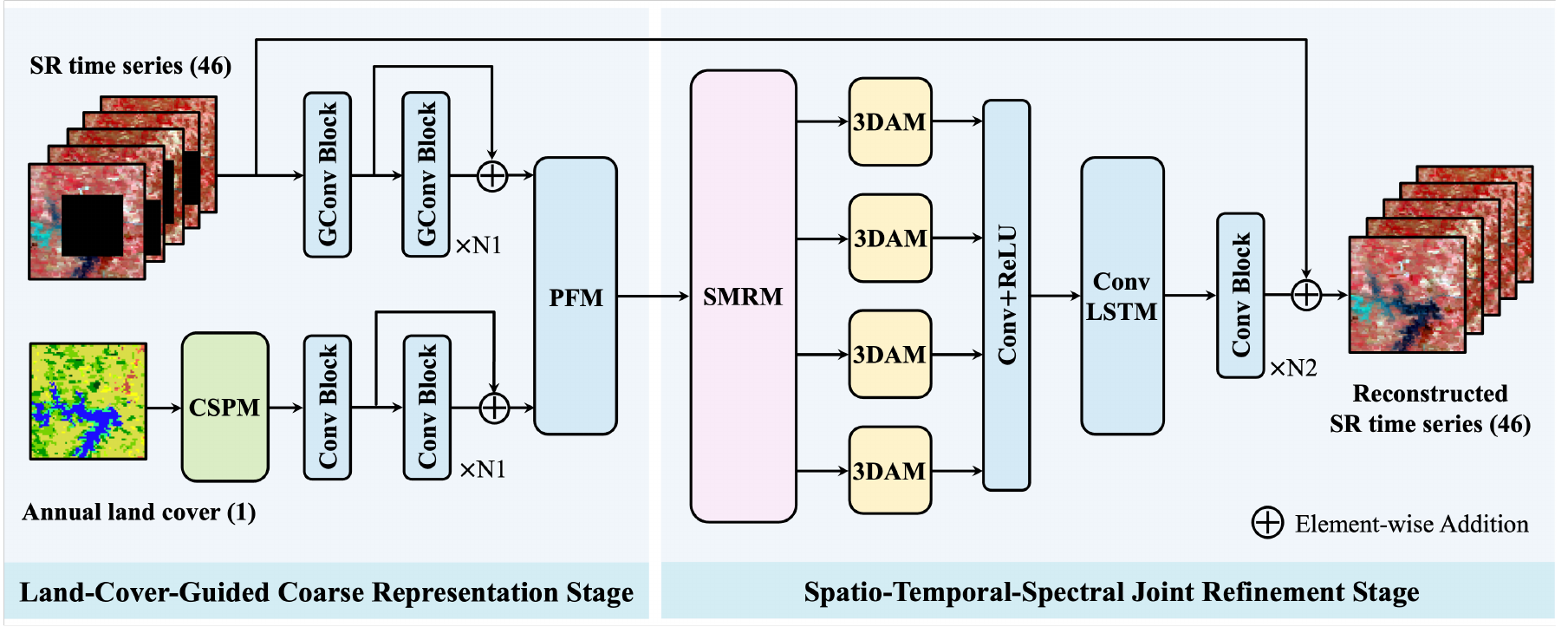}
  \caption{Overall architecture of the proposed LC-TSRNet, consisting of a land-cover-guided coarse representation stage and a spatio-temporal-spectral joint refinement stage.}
  \label{fig:4LC-TSRNet}
\end{figure*}

\begin{figure*}[htbp]
  \centering
  \includegraphics[width=\textwidth]{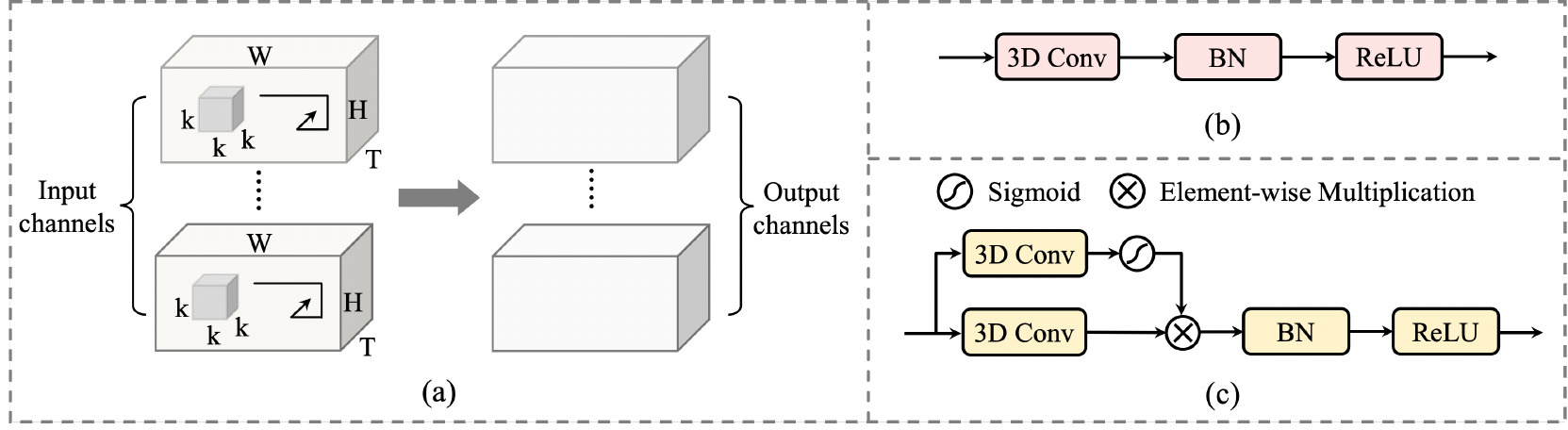}
  \caption{Illustration of (a) 3D convolution, (b) Conv block, and (c) GConv block.}
  \label{fig:53DConv}
\end{figure*}

The proposed LC-TSRNet adopts a coarse-to-fine two-stage architecture for seamless reconstruction of annual MODIS SR time series. As illustrated in Fig.~\ref{fig:4LC-TSRNet}, the network consists of two stages: a land-cover-guided coarse representation stage and a spatio-temporal-spectral joint refinement stage. Considering the temporal characteristics of MODIS data, the SR time series with missing observations is organized as an input tensor $X_{sr} \in \mathbb{R}^{C \times T \times H \times W}$, where $C$ and $T$ denote the number of channels and time steps, respectively, and $H$ and $W$ denote the image height and width. The corresponding annual land cover map is represented as $X_{lc} \in \mathbb{R}^{1 \times 1 \times H \times W}$ and input to the model. Furthermore, considering that temporal variations in SR exhibit inherent regularity and strong correlations \cite{zhu2014continuous}, all convolutional layers in LC-TSRNet employ 3D convolutions, as illustrated in Fig.~\ref{fig:53DConv}(a).

In the land-cover-guided coarse representation stage, a dual-branch architecture is designed to process the temporally incomplete SR time series and the prior reference SR, thereby generating a coarse representation. Specifically, the CSPM transforms the static land cover map into a full-year reference SR time series, providing physically consistent spectral constraints for missing regions in accordance with land cover dynamics, while standard convolutional (Conv) blocks are employed for feature extraction, as illustrated in Fig.~\ref{fig:53DConv}(b). Meanwhile, the temporally incomplete SR branch adopts gated convolution (GConv) blocks to generate adaptive gating weights and extract valid features, as illustrated in Fig.~\ref{fig:53DConv}(c). Finally, the output features from the two branches are progressively integrated by the progressive fusion module (PFM) to form an initial coarse representation.

In the spatio-temporal-spectral joint refinement stage, multiple modules are jointly employed to impose spatio-temporal-spectral constraints on the coarse representation obtained from the first stage, thereby generating seamless SR time series with improved spatiotemporal continuity. First, the SMRM adaptively exploits multi-scale information according to the extent of missing data and enables deep cross-scale interaction of spatiotemporal contextual information through a progressive aggregation mechanism. Next, the 3DAM adaptively focuses on salient features across the temporal, spatial, and spectral dimensions and imposes joint cross-dimensional constraints, thereby enhancing local textures and structural details. Then, ConvLSTM is employed to model long-range temporal dependencies and ensure temporal consistency in the reconstructed sequences. Finally, multiple cascaded Conv blocks are used to map high-dimensional features back to the original data space, while residual skip connections are introduced to preserve information in valid regions. Through the coordinated refinement of these modules, the network outputs seamless SR time series $\hat{X} \in \mathbb{R}^{C \times T \times H \times W}$ with 46 composites and 7 bands.

\subsection{Stage One: Land-Cover-Guided Coarse Representation}

In the land-cover-guided coarse representation stage, the temporally incomplete SR time series and the annual land-cover map are taken as inputs to generate a coarse reconstruction through dual-branch feature extraction and fusion. This stage mainly consists of three core components: CSPM, GConv block, and PFM. 

Specifically, the CSPM transforms land-cover semantic information into a full-year reference SR time series to provide spectral-radiometric prior constraints for missing regions. The GConv block suppresses invalid regions and extracts reliable features from the incomplete SR branch through adaptive gating. The PFM progressively integrates features from the reference SR and incomplete SR branches to form a unified coarse representation.

\subsubsection{CSPM}

To impose effective prior constraints in MODIS SR time series reconstruction, land cover information is introduced as prior knowledge. The MCD12Q1 product, as an annual land cover dataset, is well matched to the MOD09A1 SR product in spatial resolution and is readily available. Considering the significant spectral differences among different land cover classes and the relative spectral consistency within the same class \cite{small2004Landsat}, the CSPM is designed. The CSPM transforms semantic class knowledge from land cover maps into reference SR time series knowledge, thereby imposing spectral-radiometric constraints on missing regions that are consistent with the temporal dynamics of specific land cover types.

\begin{figure*}[htbp]
  \centering
  \includegraphics[width=\textwidth]{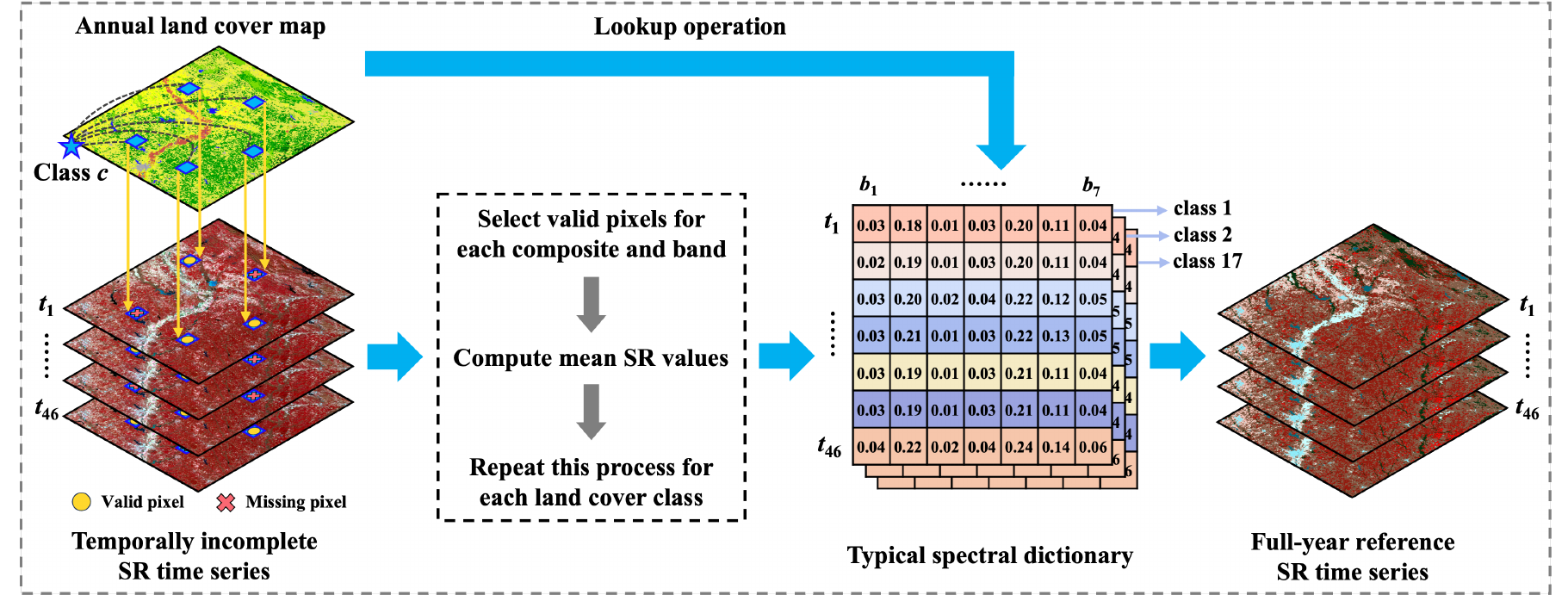}
  \caption{Schematic illustration of the CSPM.}
  \label{fig:6CSPM}
\end{figure*}

Specifically, for a given land cover class $c$ in the annual land cover map, all valid pixels (retained after QA) corresponding to this class are first selected from the incomplete MODIS SR time series. The mean SR value is then calculated for each composite and spectral band and used as the representative SR value of this class at the corresponding composite and band. By traversing all land cover classes, a typical spectral dictionary containing 17 classes, 46 time steps, and 7 bands is constructed. Finally, a lookup operation is performed on the spectral dictionary using the regional land cover map to generate a full-year reference SR time series. The schematic illustration of this process is shown in Fig.~\ref{fig:6CSPM}, and the representative SR value is calculated as follows:
\begin{equation}
    \bar{R}_{c,t,b} = \frac{1}{N_{c,t}} \sum_{i \in \Omega_{c,t}} R_{i,t,b}
    \label{eq:2spec_dict}
\end{equation}

\noindent where $\bar{R}_{c,t,b}$ denotes the mean SR of class $c$ at composite $t$ and band $b$; $R_{i,t,b}$ denotes the SR of pixel $i$ at composite $t$ and band $b$; $\Omega_{c,t}$ denotes the set of valid pixels belonging to class $c$ at composite $t$, which is shared across all bands after QA masking; and $N_{c,t}$ denotes the pixel count in $\Omega_{c,t}$.

\subsubsection{GConv Block}

Optical imagery often contains a large proportion of invalid pixels. When standard convolutions are directly applied, features from invalid regions may propagate through the network, causing noise accumulation and contaminating valid representations, which ultimately degrades reconstruction quality. To alleviate this issue, GConv blocks are introduced in the incomplete SR time series branch. Compared with the Conv block, the GConv block uses a sigmoid activation to generate explicit gating weights, thereby adaptively suppressing invalid regions and highlighting reliable pixels. This design enhances local spatiotemporal feature extraction and improves robustness under irregular gaps and persistently overlapping missing regions, as illustrated in Fig.~\ref{fig:53DConv}(c).

\subsubsection{PFM}

\begin{figure}[htbp]
  \centering
  \includegraphics[width=0.5\textwidth]{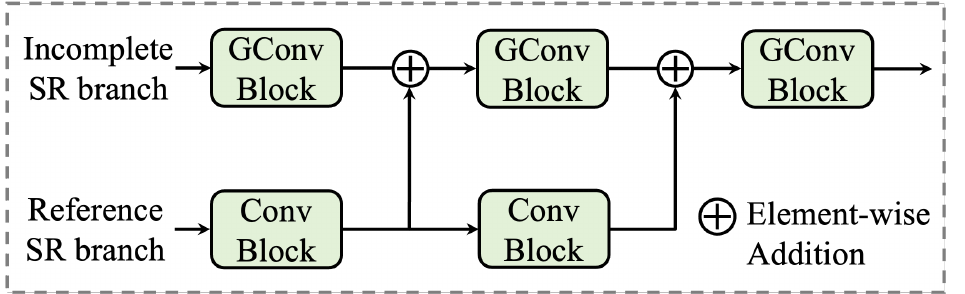}
  \caption{Illustration of the PFM structure.}
  \label{fig:7PFM}
\end{figure}

To obtain a unified coarse representation, the PFM is designed to integrate features from the dual branches, as illustrated in Fig.~\ref{fig:7PFM}. Based on a progressive fusion strategy, the PFM performs element-wise addition on multi-level feature outputs to achieve complementary information integration, effectively compensating for missing regions while preserving reliable observations. Furthermore, the PFM preserves the dual-branch design, using Conv blocks and GConv blocks to extract features from the reference SR and incomplete SR branches, respectively.

\subsection{Stage Two: Spatio-Temporal-Spectral Joint Refinement}

In the spatio-temporal-spectral joint refinement stage, the coarse representation generated in the first stage is taken as input, aiming to achieve refined reconstruction through joint spatio-temporal-spectral constraints and to generate seamless SR time series with 46 composites and 7 bands. This stage mainly consists of three core components: SMRM, 3DAM, and ConvLSTM.

Specifically, the SMRM is used to adaptively exploit multi-scale spatiotemporal contextual information, the 3DAM is employed to adaptively enhance salient features across the temporal, spatial, and spectral dimensions, and the ConvLSTM is used to model long-range temporal dependencies. Through the coordinated refinement of these modules, this stage further improves the spatiotemporal continuity, spectral fidelity, and texture detail representation of the reconstructed results.

\subsubsection{SMRM}

SR time series imagery exhibits pronounced multi-scale characteristics, where different land cover types show variability in spatial scales and temporal dynamics, while the extent of missing data is also unevenly distributed across spatiotemporal dimensions. Existing studies have demonstrated that multi-scale features play a crucial role in improving reconstruction quality \cite{xia2024crformer}, \cite{chen2017deeplab}. However, most existing methods rely on fixed local receptive fields or static multi-scale structures for feature extraction, lacking mechanisms to adaptively exploit multi-scale information according to the extent of missing data. As a result, these methods struggle to establish long-range spatiotemporal dependencies across large-area, persistent, and overlapping gaps.

To address the above issues, the SMRM is proposed in this study. To adaptively exploit multi-scale spatiotemporal contextual information based on the coarse representation obtained in the first stage, the SMRM introduces a progressive aggregation mechanism that progressively propagates and fuses the outputs of coarse-scale branches into finer-scale branches. This design enables the model to adaptively leverage contextual information at different scales according to the extent of missing data: it relies more on large-scale context in regions with extensive missing data to compensate for insufficient local information, while preserving fine-grained structural and textural details in regions with valid observations. As illustrated in Fig.~\ref{fig:8SMRM}, the SMRM consists of multiple parallel branches, each employing 3D convolutions with different dilation rates (1, 6, 12, and 18) to capture multi-scale spatiotemporal features ranging from local textures to large-scale structures. Meanwhile, each branch incorporates multiple cascaded 3D residual blocks (3DRBs) to perform deep feature extraction and representation learning. The detailed process is described as follows:

\begin{figure*}[htbp]
  \centering
  \includegraphics[width=0.8\textwidth]{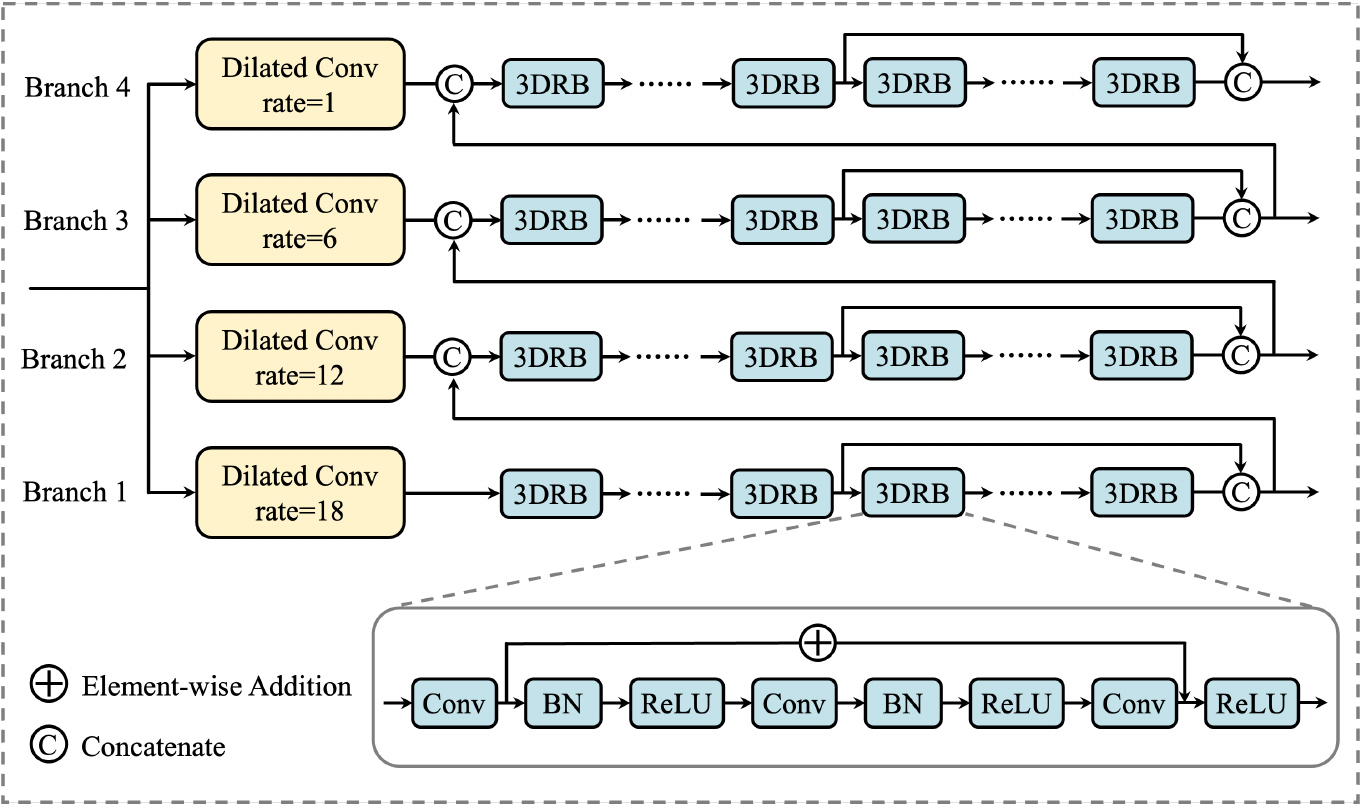}
  \caption{Structure of the SMRM.}
  \label{fig:8SMRM}
\end{figure*}

Given an input feature tensor $F \in \mathbb{R}^{C \times T \times H \times W}$, the $k$-th branch first applies a 3D dilated convolution:
\begin{equation}
F_k = \mathrm{DConv}_{r_k}(F)
\label{eq:3}
\end{equation}
\noindent where $\mathrm{DConv}_{r_k}$ denotes a 3D convolution with dilation rate $r_k$, and $k=1,2,3,4$ corresponds to dilation rates of $18$, $12$, $6$, and $1$, respectively.

To enable cross-scale interaction, a progressive aggregation mechanism is adopted, where features from coarser-scale branches are propagated to finer-scale branches:
\begin{equation}
F'_k =
\begin{cases}
F_k, & k=1, \\
\mathrm{Concat}\left(F_k, Y_{k-1}\right), & k\in\{2,3,4\}.
\end{cases}
\label{eq:4}
\end{equation}
\noindent where $\mathrm{Concat}$ denotes channel-wise concatenation, and $Y_{k-1}$ denotes the output of the preceding coarser-scale branch.

Each branch then performs deep feature extraction using multiple cascaded 3DRBs. 
Given the aggregated feature $F'_k$, the output of the $k$-th branch is obtained by concatenating intermediate and deeper representations:
\begin{equation}
Y_k = \mathrm{Concat}\left({F'_k}^{(m)}, {F'_k}^{(n)}\right)
\label{eq:5}
\end{equation}
\noindent where ${F'_k}^{(m)}$ and ${F'_k}^{(n)}$ denote the outputs of the intermediate and deeper 3DRBs in the $k$-th branch, respectively.

\subsubsection{3DAM}

\begin{figure*}[htbp]
  \centering
  \includegraphics[width=\textwidth]{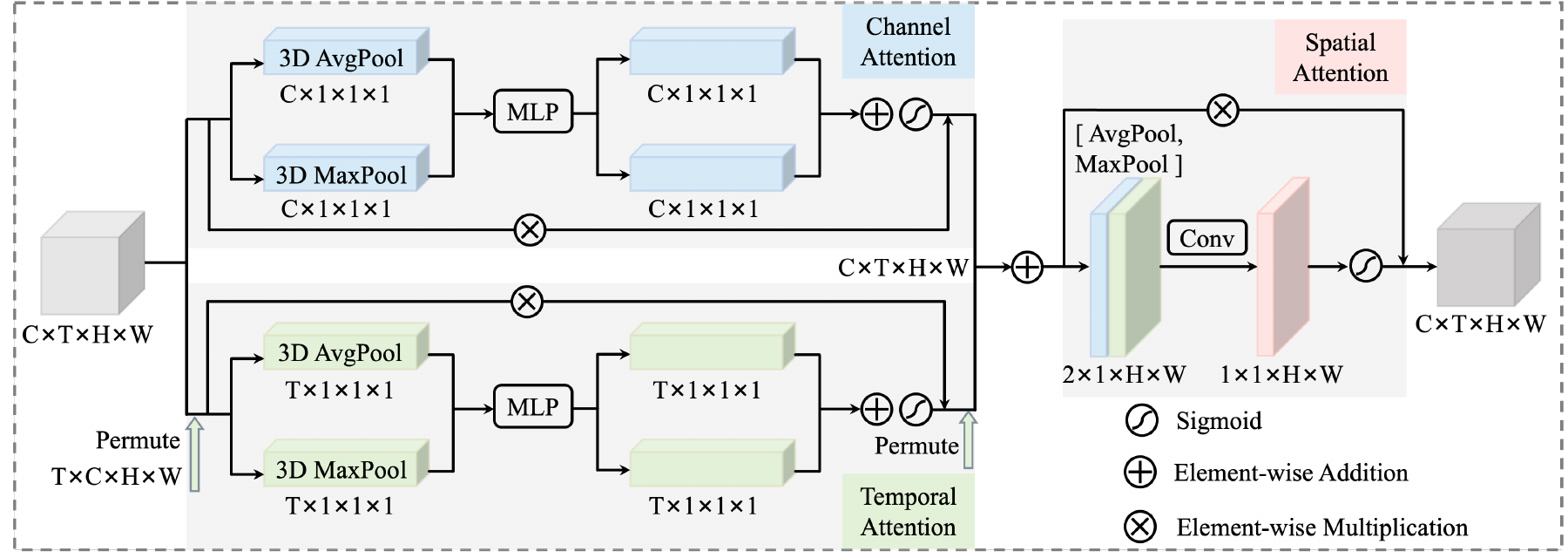}
  \caption{Structure of the 3DAM.}
  \label{fig:93DAM}
\end{figure*}

SR time series imagery contains highly coupled temporal, spatial, and spectral information, with strong intrinsic physical correlations across these dimensions \cite{jia2025spectral}. Attention mechanisms are widely used in missing reconstruction tasks to adaptively reweight feature responses and highlight informative patterns \cite{stucker2023u}, \cite{wang2025temporal}. However, existing methods typically model these dimensions independently, lacking effective joint cross-dimensional constraints. To address this issue, we develop the 3DAM inspired by CBAM \cite{woo2018cbam} and 3DACM \cite{wu2025dual}. As illustrated in Fig.~\ref{fig:93DAM}, the 3DAM consists of temporal, spatial, and channel attention submodules, which perform adaptive feature reweighting along the temporal, spatial, and spectral dimensions, respectively, and establish joint spatio-temporal-spectral constraints through a parallel-cascaded structure.

Specifically, given an input feature tensor $F \in \mathbb{R}^{C \times T \times H \times W}$, channel attention and temporal attention are first computed in parallel. For channel attention, 3D average pooling and max pooling are performed over the temporal and spatial dimensions, respectively, and the resulting descriptors are fed into a shared multilayer perceptron (MLP). The outputs of the two branches are then summed and passed through a sigmoid function to generate the channel attention map $M_c$, which is used to reweight the input features via element-wise multiplication:
\begin{gather}
M_c = \sigma\left(\mathrm{MLP}(\mathrm{AvgPool}(F)) + \mathrm{MLP}(\mathrm{MaxPool}(F))\right)
\label{eq:6}\\[5pt]
F_c = M_c \otimes F
\label{eq:7}
\end{gather}
\noindent where $F \in \mathbb{R}^{C \times T \times H \times W}$ denotes the input feature tensor; $M_c \in \mathbb{R}^{C \times 1 \times 1 \times 1}$ is the channel attention map; $F_c$ is the channel-attention-weighted feature map; $\sigma$ denotes the sigmoid activation function; and $\otimes$ denotes element-wise multiplication.

Temporal attention follows the same procedure after permuting the temporal and channel dimensions, performing pooling and the shared MLP over the channel and spatial dimensions to generate a temporal attention map $M_t$. The features are then permuted back to the original order:
\begin{gather}
\tilde{F} = \mathcal{P}_{C \leftrightarrow T}(F)
\label{eq:8}\\[5pt]
M_t = \sigma\left(\mathrm{MLP}(\mathrm{AvgPool}(\tilde{F})) + \mathrm{MLP}(\mathrm{MaxPool}(\tilde{F}))\right)
\label{eq:9}\\[5pt]
F_t = \mathcal{P}^{-1}_{C \leftrightarrow T}\left(M_t \otimes \tilde{F}\right)
\label{eq:10}
\end{gather}
\noindent where $\tilde{F} \in \mathbb{R}^{T \times C \times H \times W}$ denotes the feature map after dimension permutation; $\mathcal{P}_{C \leftrightarrow T}$ denotes the permutation operator that exchanges the channel and temporal dimensions, and $\mathcal{P}^{-1}_{C \leftrightarrow T}$ denotes its inverse; $M_t \in \mathbb{R}^{T \times 1 \times 1 \times 1}$ is the temporal attention map; and $F_t$ is the temporal-attention-weighted feature map.

Subsequently, the outputs of channel and temporal attention are combined via element-wise addition and fed into the cascaded spatial attention submodule. 
For spatial attention, average pooling and max pooling are performed over the channel and temporal dimensions, respectively. 
The pooled features are concatenated along the channel dimension, followed by a 3D convolution for feature compression. A sigmoid function is then applied to generate the spatial attention map $M_s$, which is used to further reweight the features:
\begin{gather}
F' = F_c + F_t
\label{eq:11}\\[5pt]
M_s = \sigma\left(f_{1 \times 7 \times 7}\left[\mathrm{AvgPool}(F'); \mathrm{MaxPool}(F')\right]\right)
\label{eq:12}\\[5pt]
F_s = M_s \otimes F'
\label{eq:13}
\end{gather}
\noindent where $F' \in \mathbb{R}^{C \times T \times H \times W}$ denotes the feature after channel and temporal attention reweighting and element-wise summation; $M_s \in \mathbb{R}^{1 \times 1 \times H \times W}$ is the spatial attention map; $F_s$ is the spatial-attention-weighted feature map; and $f_{1 \times 7 \times 7}$ is a 3D convolution with a kernel size of $1 \times 7 \times 7$.

\subsubsection{ConvLSTM}

ConvLSTM \cite{shi2015convolutional} is introduced in the refinement stage to further model long-range temporal dependencies. Unlike standard LSTM \cite{hochreiter1997long}, which models temporal dependencies through fully connected transformations, ConvLSTM replaces these operations with convolutional transformations, enabling temporal dependency modeling while preserving spatial structure. Therefore, ConvLSTM is well suited for MODIS SR time series reconstruction. By incorporating ConvLSTM, the model further enhances temporal consistency in the annual seamless SR time series.

\subsection{Loss Function}

To improve reconstruction accuracy, a composite loss function $L_{mr}$ is adopted for the missing reconstruction task. The loss combines a pixel-wise loss $L_{pix}$ and a structural similarity (SSIM) \cite{wang2004image} loss $L_{ssim}$ to jointly enforce numerical accuracy and structural fidelity. It is defined as follows:
\begin{equation}
    L_{mr} = L_{pix} + \lambda L_{ssim}
    \label{eq:14}
\end{equation}

\noindent where $\lambda$ is a balance factor between two loss term.

The pixel-wise loss $L_{pix}$ treats an image as a collection of pixels and measures reconstruction errors using the $L_1$ loss (MAE). To constrain reconstruction quality over both the full image and the missing regions, a global pixel loss $L_g$ and a local pixel loss $L_l$ are defined as follows:
\begin{gather}
    L_{pix} = L_g + L_l \label{eq:15} \\[5pt]
    L_g = \frac{1}{N} \left\| I - \hat{I} \right\|_1 \label{eq:16} \\[5pt]
    L_l = \frac{1}{N_m} \left\| (I - \hat{I}) \otimes (1 - M) \right\|_1 \label{eq:17}
\end{gather}

\noindent where $\hat{I}$ and $I$ denote the reconstructed and reference images, respectively; $N$ denotes the total number of pixels; $N_m$ denotes the number of pixels in the missing regions; $M$ denotes the missing mask, where 1 indicates valid pixels and 0 indicates missing pixels; and $\otimes$ denotes element-wise multiplication.

The SSIM loss $L_{ssim}$ further refines reconstruction details by constraining structural, contrast, and luminance differences between the reconstructed result and the target image. It is computed as follows:

\begin{align}
    L_{ssim} &= 1 - \mathrm{SSIM}(I, \hat{I}) \label{eq:18} \\[5pt]
    \mathrm{SSIM}(I, \hat{I}) &=
    \frac{(2\mu_I \mu_{\hat{I}} + C_1)(2\sigma_{I\hat{I}} + C_2)}
    {(\mu_I^2 + \mu_{\hat{I}}^2 + C_1)(\sigma_I^2 + \sigma_{\hat{I}}^2 + C_2)}
    \label{eq:19}
\end{align}

\noindent where $\mu_I$ and $\mu_{\hat{I}}$ denote the means of $I$ and $\hat{I}$, respectively; $\sigma_I^2$ and $\sigma_{\hat{I}}^2$ denote their variances; $\sigma_{I\hat{I}}$ denotes the covariance between $I$ and $\hat{I}$; and $C_1$ and $C_2$ are constants.

\section{Experiments and Results}
\label{Experiments}

\subsection{Experimental Settings}

\begin{figure*}[htbp]
  \centering
  \includegraphics[width=0.95\textwidth]{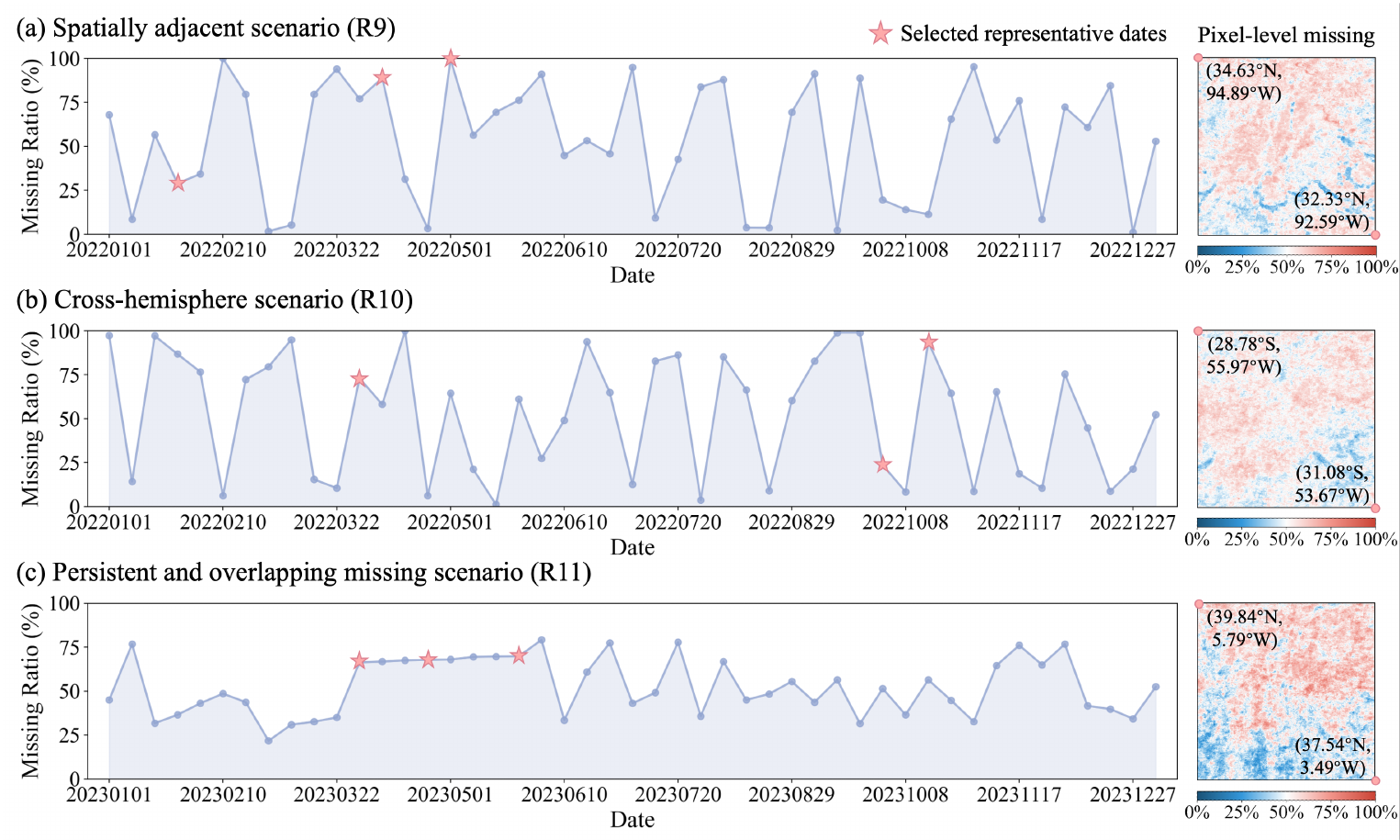}
  \caption{The regional temporal missing ratios and pixel-level annual missing ratios under different scenarios. (a) Spatially adjacent scenario (R9). (b) Cross-hemisphere scenario (R10). (c) Persistent and overlapping missing scenario (R11).}
  \label{fig:9Input}
\end{figure*}

\subsubsection{Dataset Settings}
All experiments are conducted on AIRMOD-TSR. R1--R8 are used for model training and validation, containing $4950$ samples in total. These samples are divided into training and validation sets at a ratio of $4:1$. To enhance sample diversity, data augmentation is performed during training by randomly rotating each input by $0^\circ$, $90^\circ$, $180^\circ$, or $270^\circ$.

Three high-quality ROIs are used as independent test regions under different missing scenarios. R9 represents the spatially adjacent scenario, as it is geographically close to the training ROIs. R10 represents the cross-hemisphere scenario, where seasonal phase inversion increases the difficulty of phenological reconstruction. R11 represents the persistent overlapping missing scenario for evaluating model robustness.

For all test scenarios, real MODIS missing masks obtained from GEE are applied to the high-quality ROIs to simulate realistic missing patterns. Fig.~\ref{fig:9Input} shows the regional temporal missing ratios and pixel-level annual missing ratios after mask simulation. The temporal missing ratios range from 1.14\% to 100\% for R9, from 1.13\% to 100\% for R10, and from 21.69\% to 79.11\% for R11. In R11, consecutive missing masks are further used to simulate persistent overlapping gaps. To facilitate visual comparison, three representative dates with different missing ratios and phenological stages are selected for each scenario, as indicated by the stars in Fig.~\ref{fig:9Input}.

\subsubsection{Implementation Details}

The model is implemented using PyTorch 2.5.1. Training and testing are performed on an Ubuntu 22.04 workstation equipped with NVIDIA RTX 4090 GPUs. The model is trained for 80 epochs with a batch size of 6 and an initial learning rate of $0.001$. A validation-loss-based adaptive decay schedule is adopted, where the learning rate is reduced by a factor of $0.5$ if the validation loss does not improve for four consecutive epochs. The Adam optimizer is used with $\beta_1 = 0.9$ and $\beta_2 = 0.99$. The loss balancing factor $\lambda$ is set to $0.2$.

\subsubsection{Comparison Methods}

Six representative and widely used reconstruction methods are selected for comparison, including two conventional methods, WLR \cite{zeng2013recovering} and MNSPI \cite{zhu2012modified}, and four deep learning methods, PSTCR \cite{zhang2020thick}, U-TILISE \cite{stucker2023u}, TAMGAN \cite{wang2025temporal}, and MS\textsuperscript{2}TAN \cite{zhang2025multi}. WLR is treated as a single-image-based method, whereas the remaining methods are sequence-based. For MODIS SR time series reconstruction, WLR uses the composite with the lowest missing ratio among the 46 composites as the reference image, while PSTCR divides the sequence into five-composite subsets according to acquisition time and performs group-wise reconstruction.

\subsubsection{Evaluation Metrics}

To quantitatively evaluate reconstruction performance, four widely used metrics are adopted: peak signal-to-noise ratio (PSNR) \cite{huynh2008scope}, structural similarity index measure (SSIM) \cite{wang2004image}, correlation coefficient (CC) \cite{pearson1895vii}, and spectral angle mapper (SAM) \cite{kruse1993spectral}. Specifically, PSNR and CC assess pixel-level accuracy, SAM quantifies spectral similarity by measuring the spectral angle between reconstructed and reference pixels, and SSIM evaluates structural similarity between reconstructed and reference images. In addition, root mean square error (RMSE) \cite{chai2014root} is reported as an auxiliary metric to further characterize reconstruction quality.

\subsection{Spatially Adjacent Scenario}

\begin{figure*}[htbp]
  \centering
  \includegraphics[width=\textwidth]{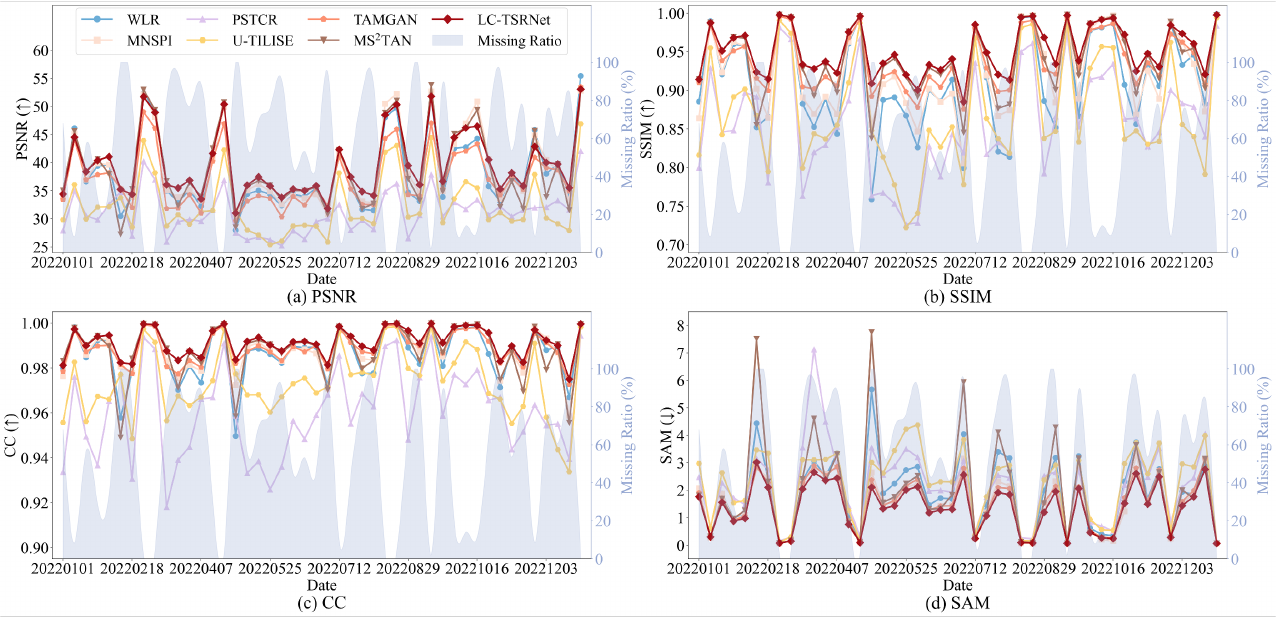}
  \caption{Time series quantitative evaluation of the reconstruction performance of different methods under the spatially adjacent scenario (R9). The shaded area represents the missing ratio for each composite. (a) PSNR. (b) SSIM. (c) CC. (d) SAM.}
  \label{fig:11R9_metrics}
\end{figure*}

In this section, the reconstruction accuracy of different methods is compared under the spatially adjacent scenario (R9), where the test ROI is geographically close to the training ROIs.

\subsubsection{Quantitative Results}

Table~\ref{tab:2R9_metrics} summarizes the quantitative evaluation of different methods on R9, with all metrics averaged across the 46 composites. The best performance is obtained by LC-TSRNet across all metrics. Compared with the second-best result for each metric, LC-TSRNet improves PSNR by 0.56 dB and reduces SAM by 8.90\%. These results indicate that LC-TSRNet achieves lower reconstruction errors, better preservation of texture structures, and higher spectral fidelity under the spatially adjacent scenario.

\begin{table}[htbp]
\centering
\renewcommand{\arraystretch}{1.2} 
\caption{Quantitative evaluation of the time series reconstruction performance of different methods under the spatially adjacent scenario (R9). Values are averaged across all 46 composites, with the best and second-best results indicated by \textbf{boldface} and \underline{underlining}, respectively.}
\label{tab:2R9_metrics}
\setlength{\tabcolsep}{6pt} 
\begin{tabular}{lcccc}
\toprule
Methods & PSNR (dB) $\uparrow$ & SSIM $\uparrow$ & CC $\uparrow$ & SAM ($^\circ$) $\downarrow$ \\
\midrule
WLR (2013) & 38.3745 & 0.9130 & 0.9865 & 1.8680 \\
MNSPI (2012) & \underline{39.1155} & 0.9251 & 0.9884 & 1.5112 \\
PSTCR (2020) & 31.1337 & 0.8593 & 0.9607 & 2.1439 \\
U-TILISE (2023) & 32.4606 & 0.8711 & 0.9737 & 2.1982 \\
TAMGAN (2025) & 37.5036 & 0.9386 & \underline{0.9891} & \underline{1.4871} \\
MS\textsuperscript{2}TAN (2025)   & 38.9208 & \underline{0.9401} & 0.9878 & 1.9728 \\
LC-TSRNet (Ours) & \textbf{39.6724} & \textbf{0.9519} & \textbf{0.9915} & \textbf{1.3548} \\
\bottomrule
\end{tabular}
\end{table}

Fig.~\ref{fig:11R9_metrics} presents the time series quantitative evaluation of the reconstruction performance on R9. Overall, the metrics track the missing ratio: higher missing ratios generally lead to lower PSNR, SSIM, and CC values, as well as higher SAM values. This indicates that the reconstruction of SR images with large-area gaps is more challenging. Among the comparison methods, PSTCR and U-TILISE show relatively poor performance on most dates, with pronounced temporal fluctuations in the evaluation metrics. This may be attributed to the sensitivity of PSTCR to the quality of multi-temporal inputs and the limited temporal modeling capability of U-TILISE. WLR, MNSPI, TAMGAN, and MS\textsuperscript{2}TAN perform well at low missing ratios but degrade markedly as the missing ratio increases, suggesting limited robustness to large-area gaps. In contrast, LC-TSRNet maintains consistently superior performance throughout the year, demonstrating stronger robustness to varying missing ratios.

\subsubsection{Qualitative Results}

\begin{figure*}[!h]
  \centering
  \includegraphics[width=\textwidth]{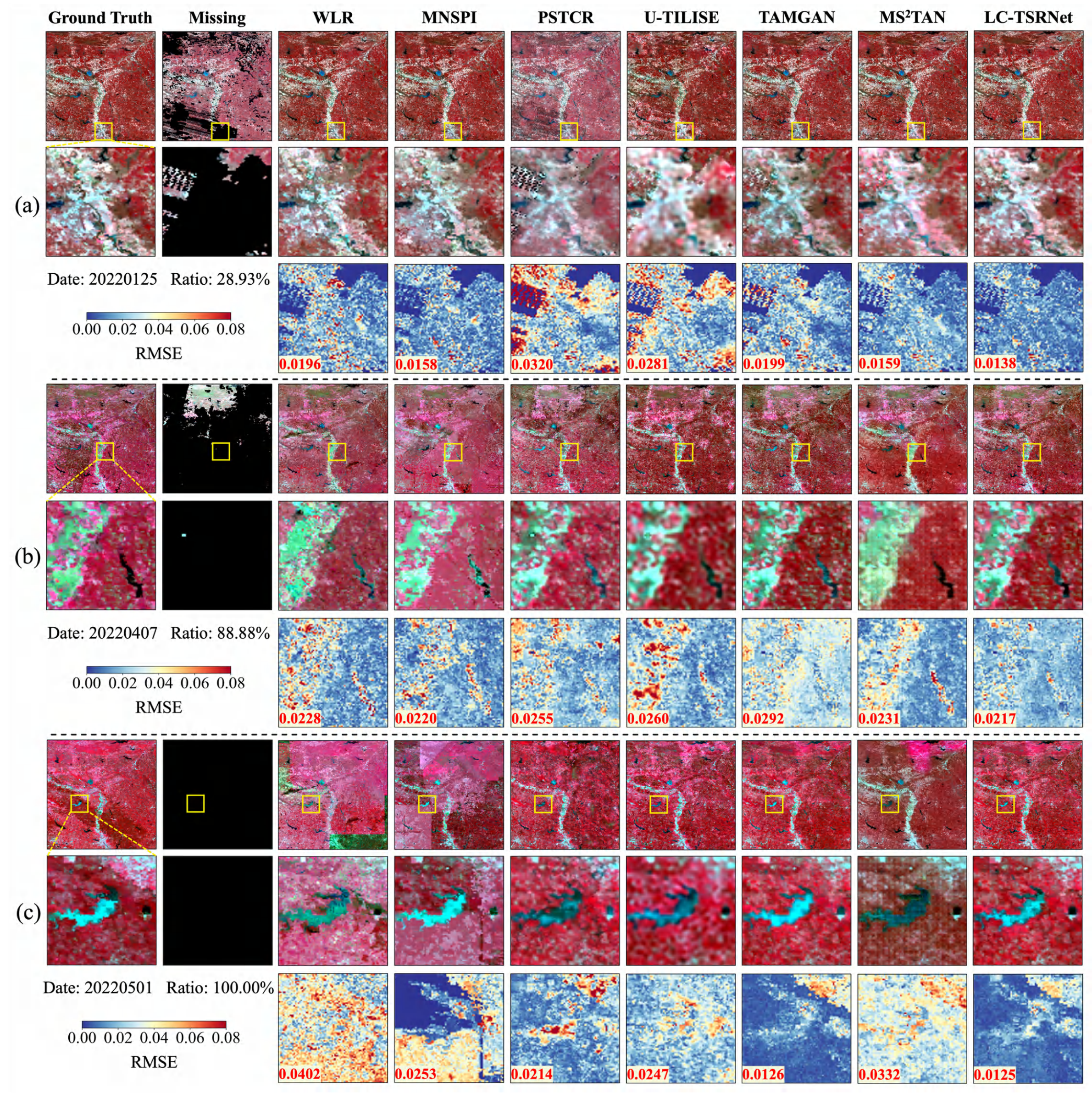}
  \caption{Visual comparison of reconstruction results on three representative dates under the spatially adjacent scenario (R9). Standard false-color composites are shown using B2, B1, and B4. For each date, the complete scene is shown in the upper row, the 80 $\times$ 80-pixel subregion marked by the yellow box is enlarged in the middle row, and the corresponding RMSE map of the enlarged region is presented in the lower row.}
  \label{fig:12R9}
\end{figure*}

Fig.~\ref{fig:12R9} presents the reconstruction results on R9 for three representative dates. The selected dates (20220125, 20220407, and 20220501) correspond to the winter stable period, rapid leaf expansion stage, and vigorous growth stage, with missing ratios of 28.93\%, 88.88\%, and 100.00\%, respectively. As shown in Fig.~\ref{fig:12R9}(a), under a lower missing ratio, satisfactory reconstruction results are obtained by most methods except PSTCR and U-TILISE. This may be attributed to the use of 2D convolution operations in PSTCR and U-TILISE, as well as their strong dependence on the quality of input data; when local information is severely insufficient, over-smoothing and loss of texture details are likely to occur. As shown in Fig.~\ref{fig:12R9}(b)--(c), under higher missing ratios, the reconstruction quality of the comparison methods decreases markedly. Large reconstruction errors and obvious color deviations are observed in WLR. The results of MNSPI suffer from over-smoothing and cross-shaped stripe artifacts, making it difficult to recover local texture details. PSTCR and U-TILISE also show texture loss and over-smoothing. TAMGAN exhibits strong generative capability, but the RMSE map still shows locally high reconstruction errors. MS\textsuperscript{2}TAN divides images into tokens for modeling based on the Vision Transformer architecture, which can easily introduce blocky artifacts. In contrast, LC-TSRNet consistently reconstructs texture details close to the reference values under different missing ratios, thereby providing the best overall visual quality.

\subsubsection{Band-Wise Reconstruction Performance}

\begin{figure*}[htbp]
  \centering
  \includegraphics[width=0.85\textwidth]{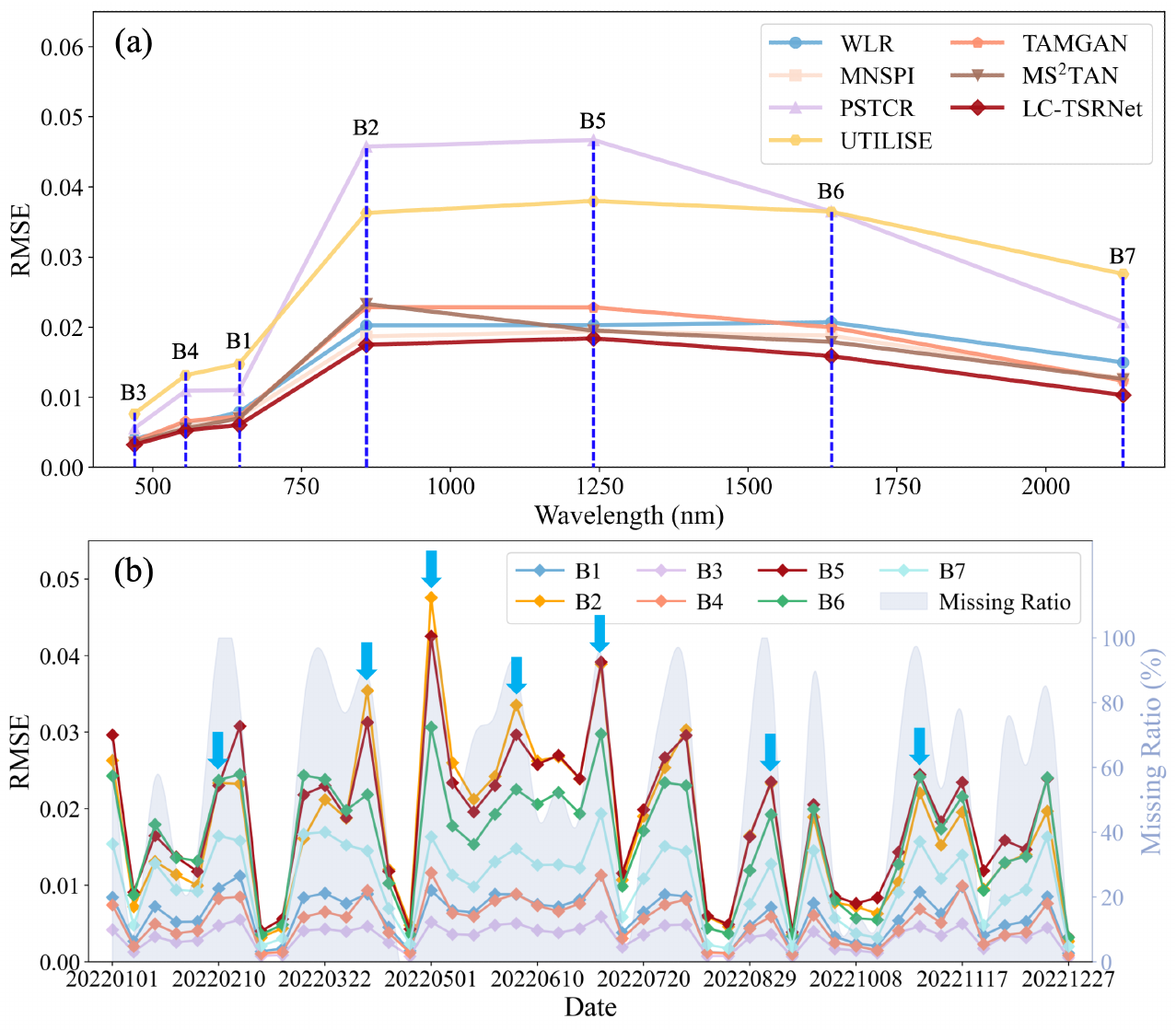}
  \caption{Band-wise RMSE analysis of reconstruction results in the spatially adjacent scenario (R9). (a) Annual mean RMSE of different methods across the seven bands. (b) Time series RMSE curves of LC-TSRNet for each band.}
  \label{fig:13_7Bands_RMSE}
\end{figure*}

To further investigate the band-wise reconstruction performance of different methods, the RMSE is calculated separately for each band of the reconstructed SR time series and averaged over all pixels. As shown in Fig.~\ref{fig:13_7Bands_RMSE}(a), LC-TSRNet achieves the lowest RMSE across all seven bands, with relatively small inter-band variation. In contrast, PSTCR and U-TILISE produce substantially higher RMSE than the other methods in all bands. A consistent band-wise pattern is observed across different methods: RMSE is relatively low in the visible bands (B3, B4, and B1), increases with wavelength, reaches its maximum in the near-infrared (NIR) band (B2), and then gradually decreases in the SWIR bands (B5, B6, and B7).

To interpret these inter-band differences, the band-wise time series RMSE curves of LC-TSRNet are further plotted in Fig.~\ref{fig:13_7Bands_RMSE}(b). According to the land cover map, vegetation accounts for 97.6\% of R9, indicating that the overall RMSE pattern is mainly dominated by reconstruction errors over vegetated pixels. As shown in Fig.~\ref{fig:13_7Bands_RMSE}(b), the RMSE of the NIR band remains relatively high and exhibits stronger temporal fluctuations throughout the year, especially during the rapid vegetation growth period from approximately April to July. This is mainly because NIR reflectance is highly sensitive to rapid changes in canopy structure and vegetation physiological conditions, which may occur within time intervals shorter than the 8-day compositing period of MOD09A1. In addition, the land cover prior derived from MCD12Q1 is updated annually and therefore cannot fully capture short-term phenological variations. When high missing ratios occur during the rapid growth period, the severe lack of valid spatial observations forces the model to rely more heavily on temporal information to infer missing regions, resulting in larger RMSE values than those observed in other periods with comparable missing ratios, as highlighted by the blue arrows in Fig.~\ref{fig:13_7Bands_RMSE}(b).

\subsubsection{Spectral Fidelity Analysis of Typical Land Cover Types}

\begin{figure*}[htbp]
  \centering
  \includegraphics[width=\textwidth]{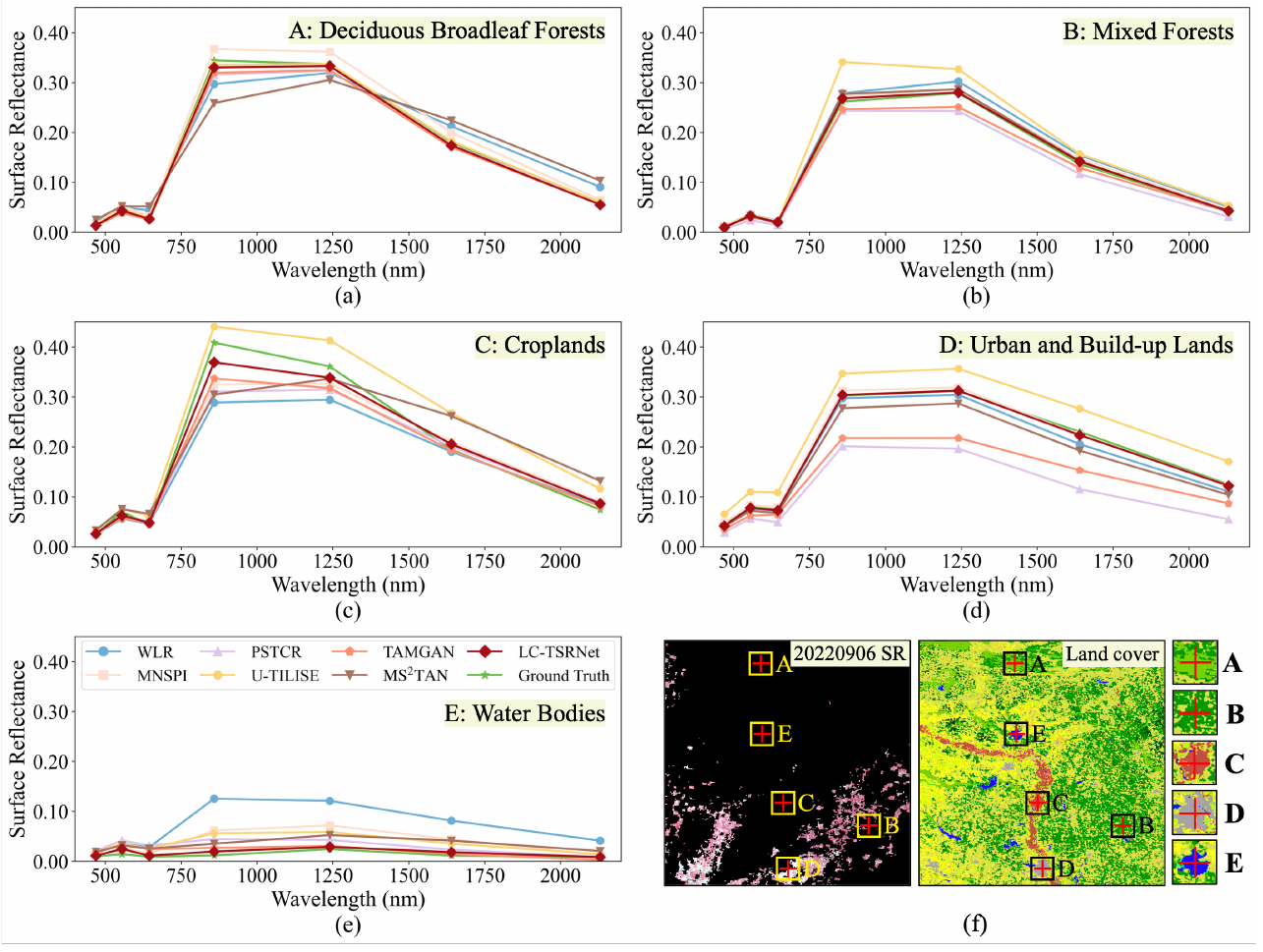}
  \caption{Comparison of reconstructed SR curves for typical land cover types derived from the R9 image acquired on 20220906. The spatial locations of selected samples are also shown. (a) Deciduous broadleaf forests. (b) Mixed forests. (c) Croplands. (d) Urban and built-up lands. (e) Water bodies. (f) Spatial locations of selected samples.}
  \label{fig:14SpeCurve}
\end{figure*}

The test ROI covers a large spatial extent and includes diverse land cover types. To further evaluate the spectral fidelity of the reconstructed imagery, SR curve analysis of typical land cover types is conducted.

First, the image acquired on 20220906 is selected to compare the SR curves of five typical land cover types, including deciduous broadleaf forests, mixed forests, croplands, urban and built-up lands, and water bodies, as shown in Fig.~\ref{fig:14SpeCurve}. Overall, the spectral curves produced by LC-TSRNet are closest to the original observations across all five land cover types, indicating its excellent spectral fidelity. Specifically, in deciduous broadleaf forests, MS\textsuperscript{2}TAN shows obvious underestimation in the NIR band. In mixed forests, U-TILISE exhibits clear overestimation in the NIR band. In croplands, relatively large spectral deviations are produced by all comparison methods in the NIR band, suggesting that spectral reconstruction in densely vegetated areas remains challenging. In urban and built-up lands, obvious underestimation is observed in the infrared bands for PSTCR and TAMGAN, whereas U-TILISE shows clear overestimation. In water bodies, WLR tends to overestimate reflectance in the infrared bands.

\begin{figure*}[!h]
  \centering
  \includegraphics[width=0.9\textwidth]{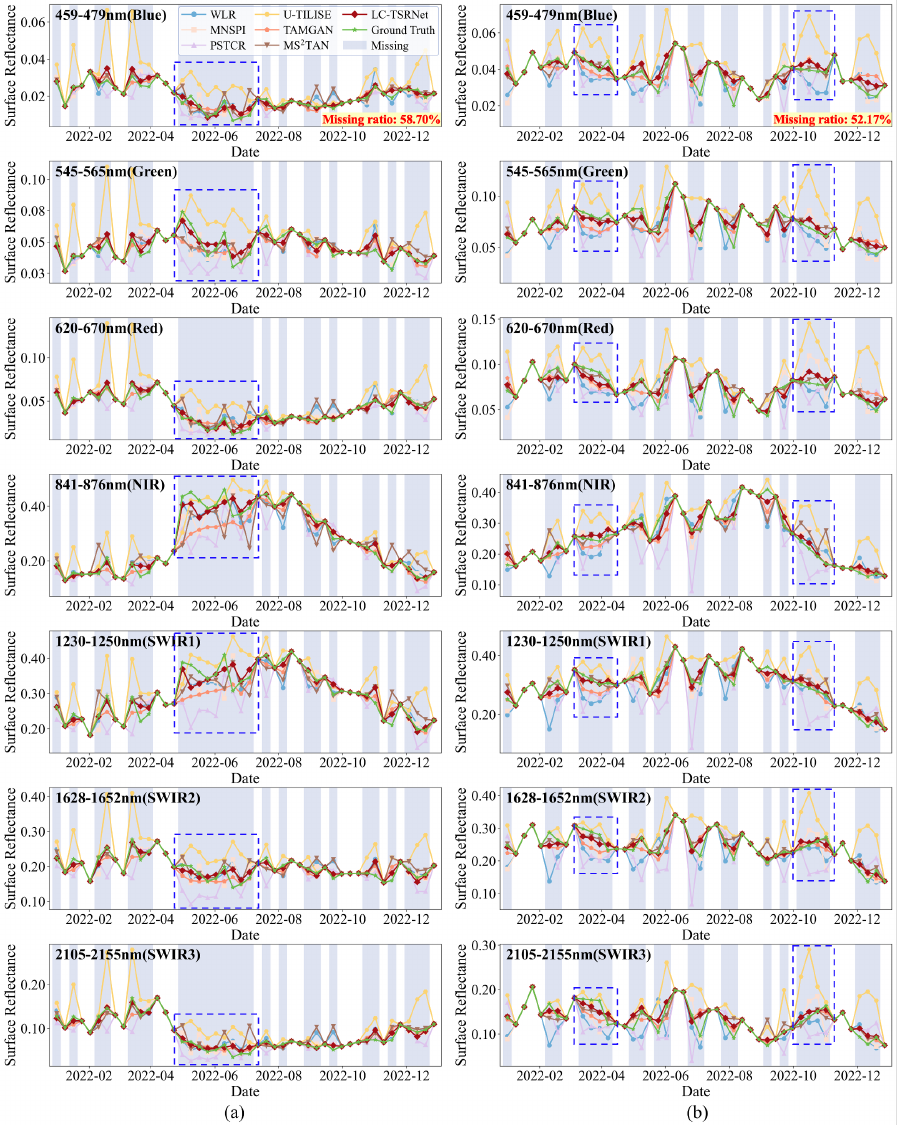}
  \caption{Comparison of reconstructed time series SR curves over vegetated areas. Blue shaded areas indicate dates with missing observations. (a) Deciduous broadleaf forests. (b) Croplands.}
  \label{fig:15SpeCurveTS}
\end{figure*}

In addition, for vegetation types with strong temporal dynamics, SR variations are largely driven by phenological processes and therefore exhibit pronounced seasonal changes. Fig.~\ref{fig:15SpeCurveTS} compares the reconstructed time series SR curves of two representative vegetation samples, including deciduous broadleaf forests and croplands, with temporal missing ratios of 58.70\% and 52.17\%, respectively. Under short-term missing conditions, most methods can generally follow the seasonal variation of the ground truth. However, under persistent long-term gaps, such as the continuous missing period from May to July for deciduous broadleaf forests, larger spectral deviations and stronger temporal fluctuations are observed in the comparison methods. In contrast, the SR curves reconstructed by LC-TSRNet are closer to the ground truth throughout the year, indicating stronger robustness to persistent missing observations and better preservation of key phenological patterns.

\subsection{Cross-Hemisphere Scenario}

In this section, the generalization capability of different methods is evaluated in the cross-hemisphere scenario (R10), where the test ROI is geographically distant from the training ROIs.

\subsubsection{Quantitative Results}

Table~\ref{tab:3R10_metrics} summarizes the quantitative evaluation of different methods on R10, with all metrics averaged across the 46 composites. Compared with R9, performance degradation is observed for all methods across all metrics, reflecting the increased reconstruction difficulty caused by substantial geographic and seasonal differences between the training and test ROIs. Nevertheless, the best performance is still obtained by LC-TSRNet across all metrics. Compared with the second-best result for each metric, LC-TSRNet improves PSNR by 0.32 dB and reduces SAM by 6.20\%, indicating reduced spectral bias under the cross-hemisphere scenario.

\begin{table}[htbp]
\centering
\renewcommand{\arraystretch}{1.2} 
\caption{Quantitative evaluation of the time series reconstruction performance of different methods under the cross-hemisphere scenario (R10). Values are averaged across all 46 composites, with the best and second-best results indicated by \textbf{boldface} and \underline{underlining}, respectively.}
\label{tab:3R10_metrics}
\setlength{\tabcolsep}{6pt} 
\begin{tabular}{lcccc}
\toprule
Methods & PSNR (dB) $\uparrow$ & SSIM $\uparrow$ & CC $\uparrow$ & SAM ($^\circ$) $\downarrow$ \\
\midrule
WLR (2013) & 35.8984 & 0.8808 & 0.9821 & 2.1624 \\
MNSPI (2012) & \underline{37.0999} & 0.9097 & \underline{0.9867} & \underline{1.5321} \\
PSTCR (2020) & 28.8164 & 0.8243 & 0.9526 & 2.2271 \\
U-TILISE (2023) & 32.4131 & 0.8670 & 0.9776 & 2.1690 \\
TAMGAN (2025) & 34.9539 & \underline{0.9196} & 0.9864 & 1.5806 \\
MS\textsuperscript{2}TAN (2025)  & 35.8663 & 0.9050 & 0.9840 & 2.1591 \\
LC-TSRNet (Ours) & \textbf{37.4161} & \textbf{0.9348} & \textbf{0.9894} & \textbf{1.4371} \\
\bottomrule
\end{tabular}
\end{table}

\begin{figure*}[htbp]
  \centering
  \includegraphics[width=\textwidth]{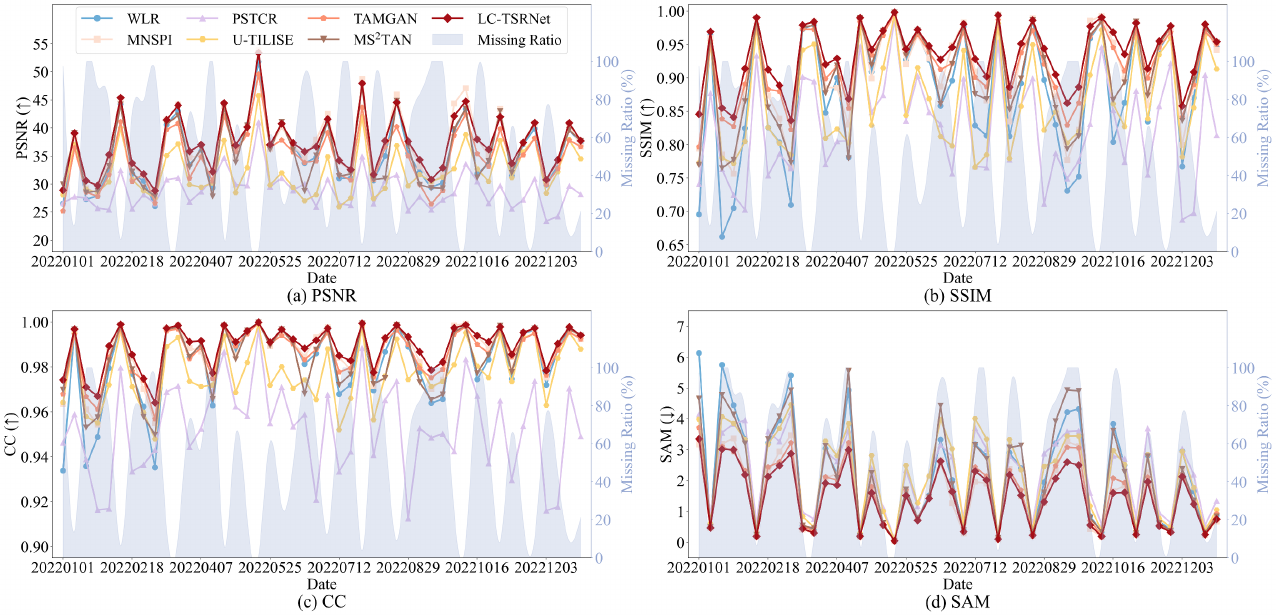}
  \caption{Time series quantitative evaluation of the reconstruction performance of different methods under the cross-hemisphere scenario (R10). The shaded area represents the missing ratio for each composite. (a) PSNR. (b) SSIM. (c) CC. (d) SAM.}
  \label{fig:16R10_metrics}
\end{figure*}

Fig.~\ref{fig:16R10_metrics} presents the time series quantitative evaluation of the reconstruction performance on R10. PSTCR and U-TILISE generally perform worse than the other methods and exhibit larger temporal fluctuations in the evaluation metrics, indicating limited generalization under the cross-hemisphere scenario. On several dates with high missing ratios, WLR and MS\textsuperscript{2}TAN show sharp decreases in SSIM and increases in SAM, suggesting insufficient robustness to severe observation gaps. In contrast, LC-TSRNet maintains stable and accurate reconstruction performance across the time series, even under high missing ratios, demonstrating strong generalization to heterogeneous cross-hemisphere conditions.

\subsubsection{Qualitative Results}

\begin{figure*}[!h]
  \centering
  \includegraphics[width=\textwidth]{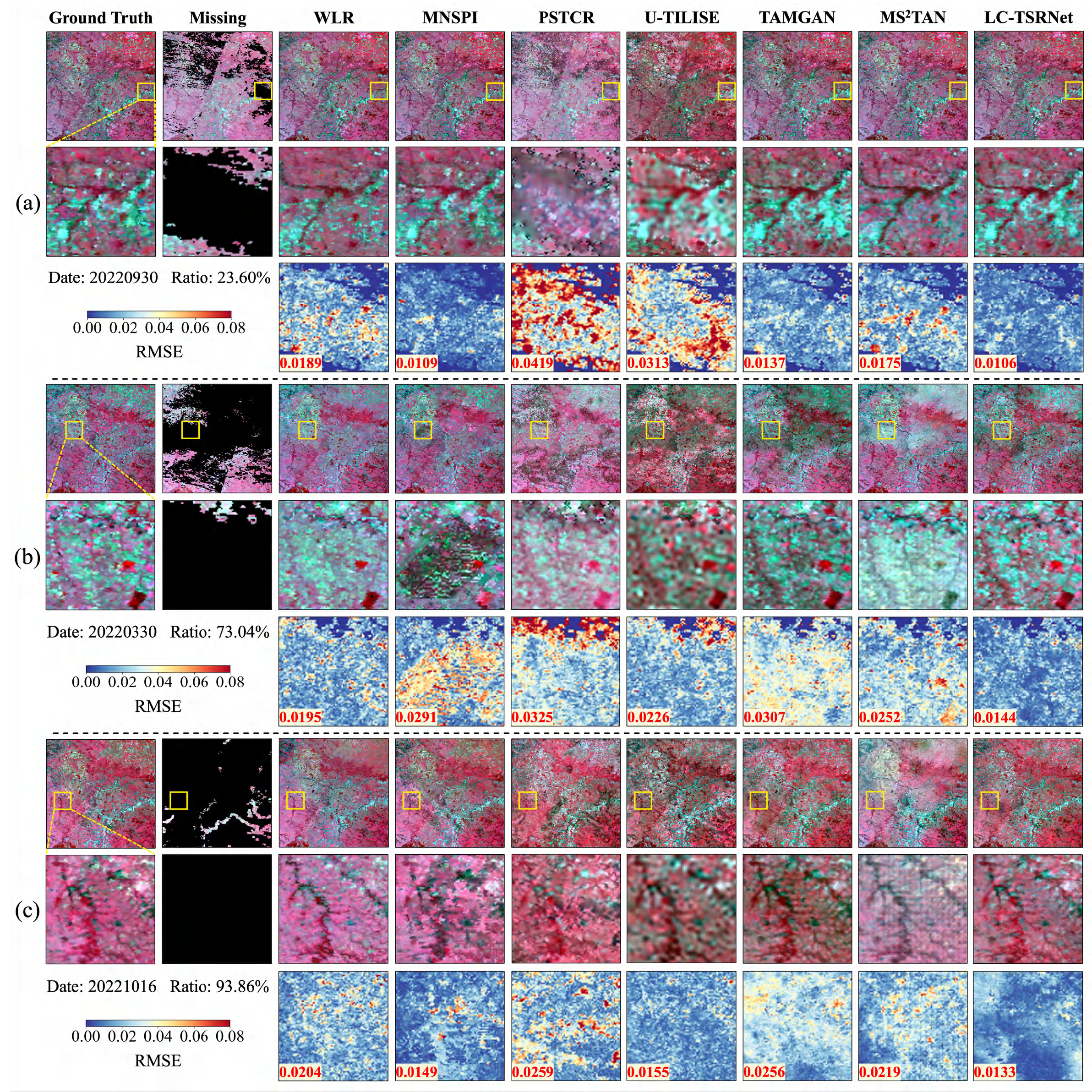}
  \caption{Visual comparison of reconstruction results on three representative dates under the cross-hemisphere scenario (R10). Standard false-color composites are shown using B2, B1, and B4. For each date, the complete scene is shown in the upper row, the 80 $\times$ 80-pixel subregion marked by the yellow box is enlarged in the middle row, and the corresponding RMSE map of the enlarged region is presented in the lower row.}
  \label{fig:17R10}
\end{figure*}

Fig.~\ref{fig:17R10} presents the reconstruction results on R10 for three representative dates. The selected dates (20220930, 20220330, and 20221016) correspond to the early growth stage, vegetation senescence stage, and rapid growth stage, with missing ratios of 23.60\%, 73.04\%, and 93.86\%, respectively. Similar to the reconstruction performance on R9, satisfactory visual reconstruction results are obtained by most methods under the lower missing ratio, except that PSTCR and WLR exhibit a certain degree of over-smoothing, as shown in Fig.~\ref{fig:17R10}(a). As the missing ratio increases, the reconstruction quality of the comparison methods decreases noticeably. Due to the Laplacian-prior regularization adopted by WLR, texture structure loss is likely to occur in regions with insufficient available auxiliary pixels, and this phenomenon is particularly evident in the bottom region of the image on 20221016. As shown in Fig.~\ref{fig:17R10}(b)--(c), severe artifacts and texture blurring are observed in the MNSPI results, making it difficult to recover local details effectively. PSTCR and WLR also show obvious texture loss and blurring. Although TAMGAN has strong generative capability, color distortion still remains. MS\textsuperscript{2}TAN continues to exhibit blocky artifacts under high missing ratios. In contrast, LC-TSRNet robustly restores texture details and color information close to the reference images across different missing ratios, while maintaining consistently low RMSE values, indicating strong generalization capability in the cross-hemisphere scenario.

\begin{table}[htbp]
\centering
\renewcommand{\arraystretch}{1.2} 
\caption{Quantitative evaluation of the time series reconstruction performance of different methods under the persistent and overlapping missing scenario (R11). Values are averaged across 8 composites, with the best and second-best results indicated by \textbf{boldface} and \underline{underlining}, respectively.}
\label{tab:4R11_metrics}
\setlength{\tabcolsep}{6pt} 
\begin{tabular}{lcccc}
\toprule
Methods & PSNR (dB) $\uparrow$ & SSIM $\uparrow$ & CC $\uparrow$ & SAM ($^\circ$) $\downarrow$ \\
\midrule
WLR (2013) & 30.5794 & 0.8541 & 0.9586 & 1.9029 \\
MNSPI (2012) & \underline{32.1557} & 0.8524 & \underline{0.9757} & 1.9190 \\
PSTCR (2020) & 24.4754 & 0.6779 & 0.8886 & 3.2217 \\
U-TILISE (2023) & 28.0706 & 0.7981 & 0.9528 & 2.9854 \\
TAMGAN (2025) & 30.8160 & 0.8734 & 0.9745 & 1.9762 \\
MS\textsuperscript{2}TAN (2025) & 32.0039 & \underline{0.8875} & 0.9749 & \underline{1.8996} \\
LC-TSRNet (Ours) & \textbf{32.3261} & \textbf{0.8931} & \textbf{0.9774} & \textbf{1.8807} \\
\bottomrule
\end{tabular}
\end{table}

\subsection{Persistent and Overlapping Missing Scenario}

In this section, the robustness of different methods is assessed under the persistent overlapping missing scenario (R11). With an average missing ratio of 52.47\% and a minimum missing ratio of 21.69\%, R11 contains no relatively complete high-quality images, making it a challenging case for gap filling.

\begin{figure*}[!h]
  \centering
  \includegraphics[width=\textwidth]{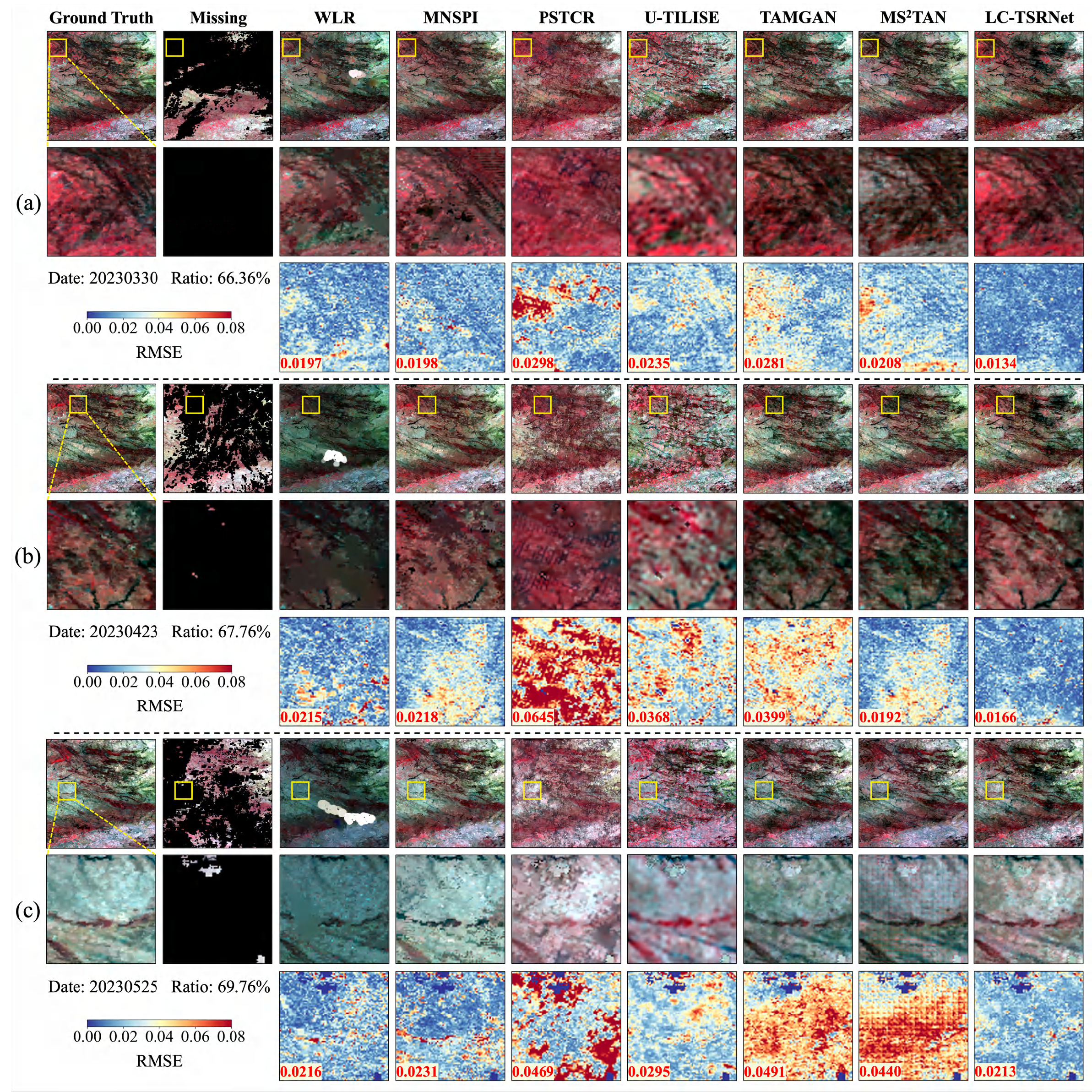}
  \caption{Visual comparison of reconstruction results on three representative dates under the persistent and overlapping missing scenario (R11). Standard false-color composites are shown using B2, B1, and B4. For each date, the complete scene is shown in the upper row, the 80 $\times$ 80-pixel subregion marked by the yellow box is enlarged in the middle row, and the corresponding RMSE map of the enlarged region is presented in the lower row.}
  \label{fig:18R11}
\end{figure*}

\subsubsection{Quantitative Results}

Table~\ref{tab:4R11_metrics} summarizes the quantitative evaluation of different methods on R11, with all metrics averaged over eight consecutive composites from 20230330 to 20230525. The missing ratios of these composites range from 66.36\% to 69.76\%, representing a persistent overlapping missing scenario. Compared with R9 and R10, all methods exhibit further degradation across the evaluation metrics, demonstrating the greater reconstruction difficulty caused by continuous large-area missing observations. Despite this challenging condition, the best performance is still obtained by LC-TSRNet across all metrics.

\subsubsection{Qualitative Results}

Fig.~\ref{fig:18R11} presents the reconstruction results on R11 for three representative dates. The selected dates (20230330, 20230423, and 20230525) correspond to the transition from the vigorous vegetation growth stage to the early senescence stage. Due to the absence of relatively complete reference images, WLR mainly relies on Laplacian-prior regularization to fill regions with limited reference pixels, leading to obvious over-smoothing and texture loss. MNSPI, PSTCR, and U-TILISE also suffer from severe texture loss and relatively large RMSE values. TAMGAN and MS\textsuperscript{2}TAN exhibit evident color bias and large reconstruction errors, particularly on 20230423 and 20230525. In contrast, LC-TSRNet preserves clearer textures and more consistent color information, demonstrating stronger robustness under persistent overlapping gaps.

\subsection{Ablation studies}\label{Ablation}

\begin{table*}[htbp] 
\centering
\renewcommand{\arraystretch}{1.2} 
\caption{Quantitative ablation study of LC-TSRNet under the spatially adjacent scenario (R9) and cross-hemisphere scenario (R10). Values are averaged across all 46 composites, with the best results indicated by \textbf{boldface}.}
\label{tab5:ablation}
\setlength{\tabcolsep}{6pt} 
\begin{tabular}{cccccccccccc}
\toprule
\multicolumn{4}{c}{Methods} & \multicolumn{4}{c}{R9} & \multicolumn{4}{c}{R10} \\
\cmidrule(lr){1-4} \cmidrule(lr){5-8} \cmidrule(lr){9-12}
Land Cover & CSPM & SMRM & 3DAM & PSNR (dB) $\uparrow$ & SSIM $\uparrow$ & CC $\uparrow$ & SAM ($^\circ$) $\downarrow$ & PSNR (dB) $\uparrow$ & SSIM $\uparrow$ & CC $\uparrow$ & SAM ($^\circ$) $\downarrow$ \\
\midrule

$\times$ & $\times$ & \checkmark & \checkmark & 38.6871 & 0.9465 & 0.9901 & 1.4707 & 31.2672 & 0.8976 & 0.9582 & 2.2841 \\

\checkmark & $\times$ & \checkmark & \checkmark & 39.0090 & 0.9490 & 0.9909 & 1.4496 & 33.3313 & 0.9079 & 0.9828 & 1.9829 \\

\checkmark & \checkmark & $\times$ & \checkmark & 39.4310 & 0.9474 & 0.9912 & 1.4902 & 37.1209 & 0.9310 & 0.9887 & 1.5843 \\

\checkmark & \checkmark & \checkmark & $\times$ & 38.9883 & 0.9456 & 0.9906 & 1.4884 & 37.0544 & 0.9331 & 0.9889 & 1.5165 \\

\checkmark & \checkmark & \checkmark & \checkmark & \textbf{39.6724} & \textbf{0.9519} & \textbf{0.9915} & \textbf{1.3548} & \textbf{37.4161} & \textbf{0.9348} & \textbf{0.9894} & \textbf{1.4371} \\

\bottomrule
\end{tabular}
\end{table*}

\begin{figure*}[htbp]
  \centering
  \includegraphics[width=0.75\textwidth]{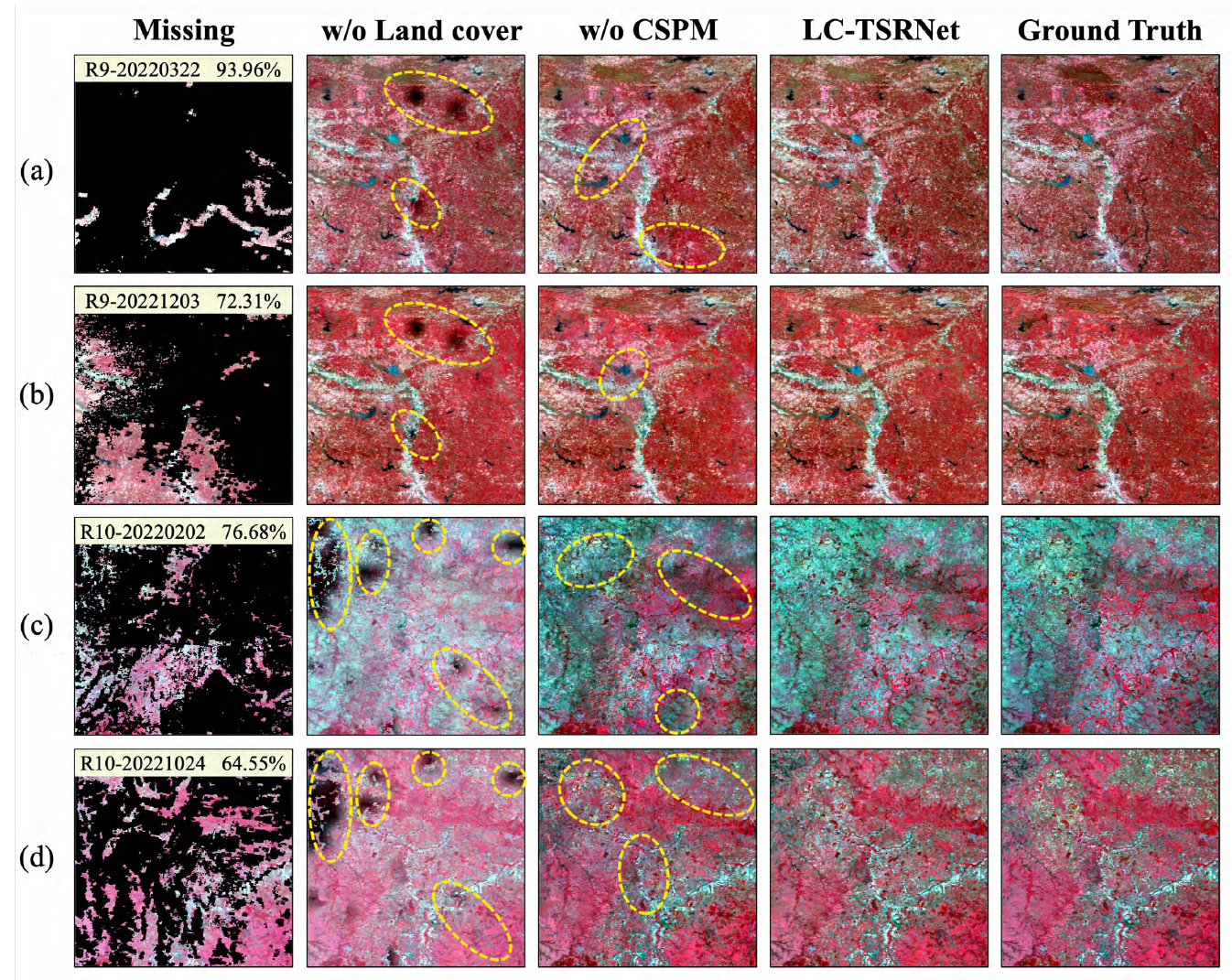}
  \caption{Visual comparison of ablation results for the land cover prior and CSPM in the spatially adjacent scenario (R9) and cross-hemisphere scenario (R10). (a) R9-20220322. (b) R9-20221203. (c) R10-20220202. (d) R10-20221024.}
  \label{fig:19Abl}
\end{figure*}

To investigate the contributions of individual modules and the land cover prior, ablation studies are conducted under the spatially adjacent scenario (R9) and the cross-hemisphere scenario (R10). When the land cover prior is removed, only incomplete SR time series are used as input. When CSPM is removed, land cover class labels are directly used as input. When SMRM is removed, only a single branch with a dilation rate of 1 is retained. When 3DAM is removed, the multi-scale representations are first refined by a convolution unit and then fed into ConvLSTM to model long-term temporal dependencies.

\subsubsection{Quantitative Results}

Table~\ref{tab5:ablation} summarizes the ablation results of five settings: removing the land cover prior, CSPM, SMRM, and 3DAM, and using the complete LC-TSRNet. The results show that removing any component degrades the reconstruction performance on both R9 and R10, confirming the effectiveness of each component in the proposed framework. Among them, the land cover prior and CSPM provide particularly notable improvements. For the geographically distant R10 region, introducing the land cover prior improves PSNR by 6.60\% and reduces SAM by 13.19\% compared with using only incomplete SR time series. Furthermore, compared with directly using land cover labels, CSPM further improves PSNR by 12.26\% and reduces SAM by 27.53\%. These results indicate that land cover information enhances cross-region reconstruction, while CSPM further improves spectral fidelity by converting semantic labels into spectral priors.

\subsubsection{Qualitative Results}

Fig.~\ref{fig:19Abl} presents the ablation results of the land cover prior and CSPM on R9 and R10. When the land cover prior is removed, noticeable black speckle artifacts appear in the reconstructed images. This is mainly because the model relies only on the incomplete SR branch, where gated convolutions may over-suppress feature responses in severely missing regions, leading to localized reconstruction failures. When CSPM is removed, distinct color deviations and residual noise are observed in several regions, indicating that directly using land cover labels cannot provide sufficient spectral constraints. These results further demonstrate the importance of land cover guidance and class-specific spectral priors for stable and spectrally consistent reconstruction.

\section{Discussion}
\label{Discussion}

\subsection{Reconstruction Performance under Real-world Scenario}

\begin{figure*}[htbp]
  \centering
  \includegraphics[width=0.9\textwidth]{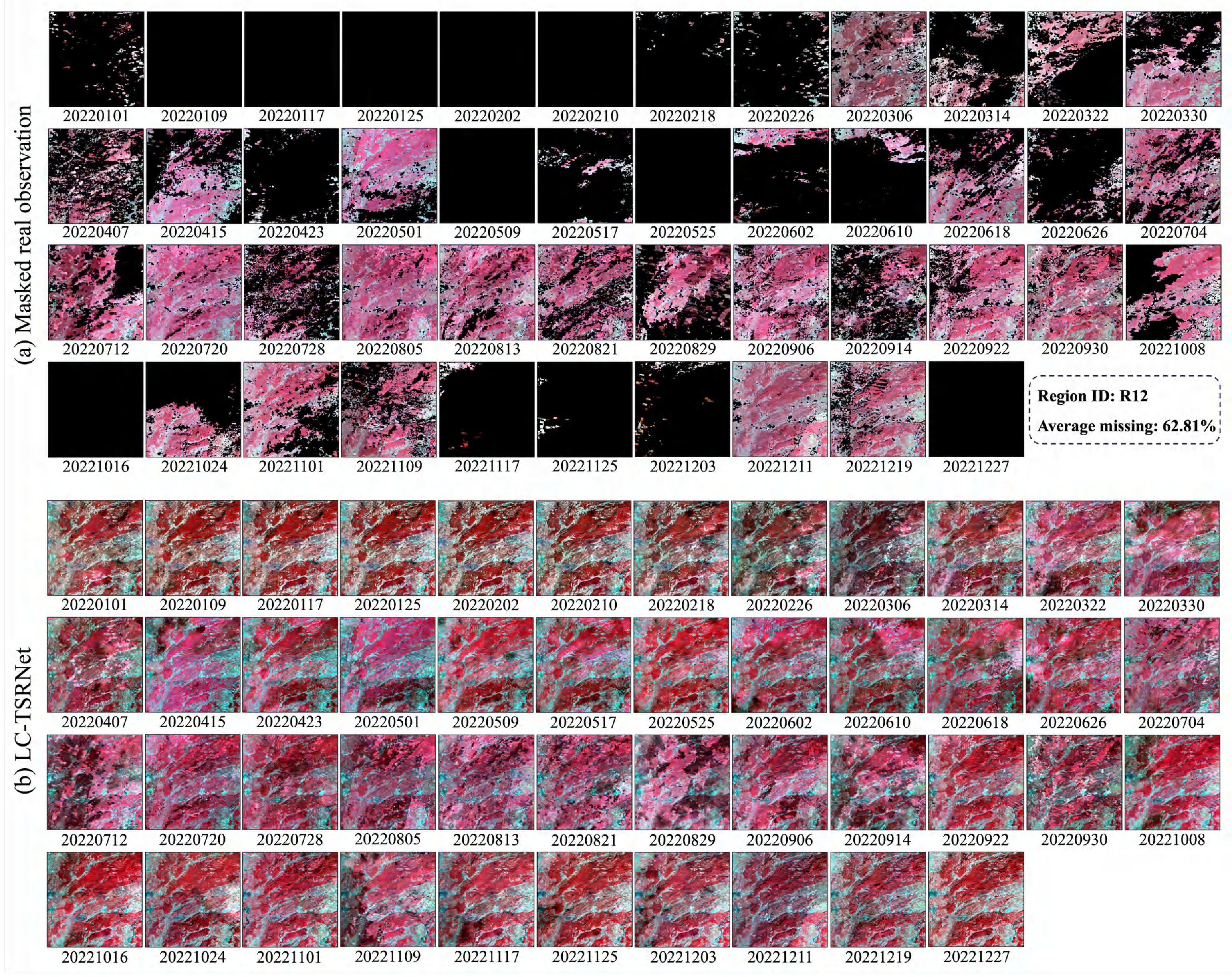}
  \caption{Visual comparison of masked real observation time series and LC-TSRNet reconstruction results in the real-world scenario (R12). (a) Masked real observation time series. (b) Reconstruction results of LC-TSRNet.}
  \label{fig:20R12_Real}
\end{figure*}

To evaluate reconstruction performance under real-world observation conditions, R12 is selected as the ROI for real-data experiments. In 2022, the average missing ratio of the 46 composites in R12 reaches 62.81\%, with several dates approaching complete missingness.

Fig.~\ref{fig:20R12_Real} shows the masked real observed SR time series on R12 and the corresponding reconstruction results of LC-TSRNet. The real observations exhibit a distinct temporally uneven missing patterns, with relatively complete observations on some dates and severe data gaps on others. In particular, large-area gaps persist for nearly two consecutive months from 20220101 to 20220226, posing a substantial challenge to the reconstruction capability of the model. Nevertheless, LC-TSRNet is still able to reconstruct spatially continuous images, with smooth transitions between reconstructed regions and originally valid observations and without obvious boundary artifacts. In addition, the reconstructed images on adjacent dates maintain good consistency in texture structures and color information. For other severely missing dates, such as 20220509, 20220525, 20221016, and 20221227, LC-TSRNet also achieves satisfactory reconstruction results. These results indicate that LC-TSRNet has strong robustness and practical applicability under real-world observation conditions, providing an effective solution for generating seamless MOD09A1 SR product from incomplete observations.

\subsection{Applications in Vegetation Monitoring}

\begin{figure*}[!h]
  \centering
  \includegraphics[width=0.9\textwidth]{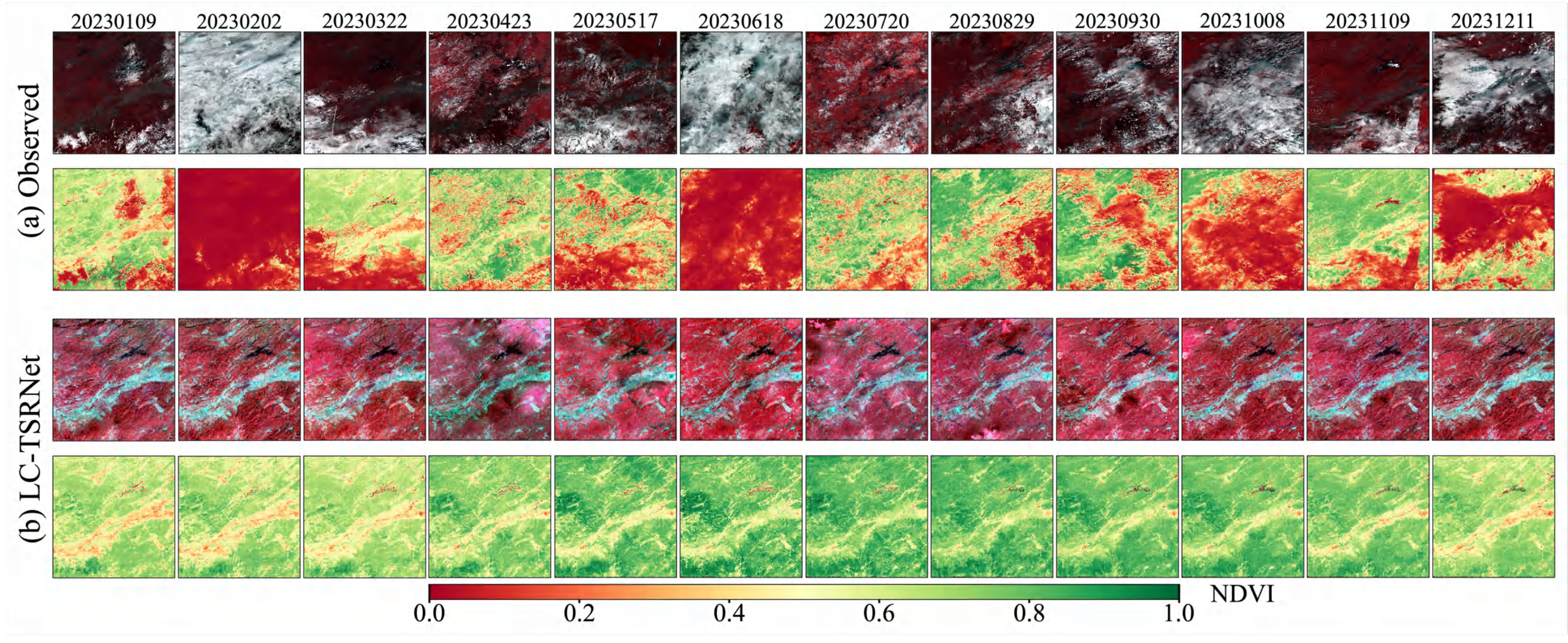}
  \caption{Visual comparison of observed and reconstructed SR and NDVI time series in the application region (R13). (a) Observed time series. (b) Reconstruction results of LC-TSRNet.}
  \label{fig:21NDVI}
\end{figure*}

\begin{figure*}[!h]
  \centering
  \includegraphics[width=0.8\textwidth]{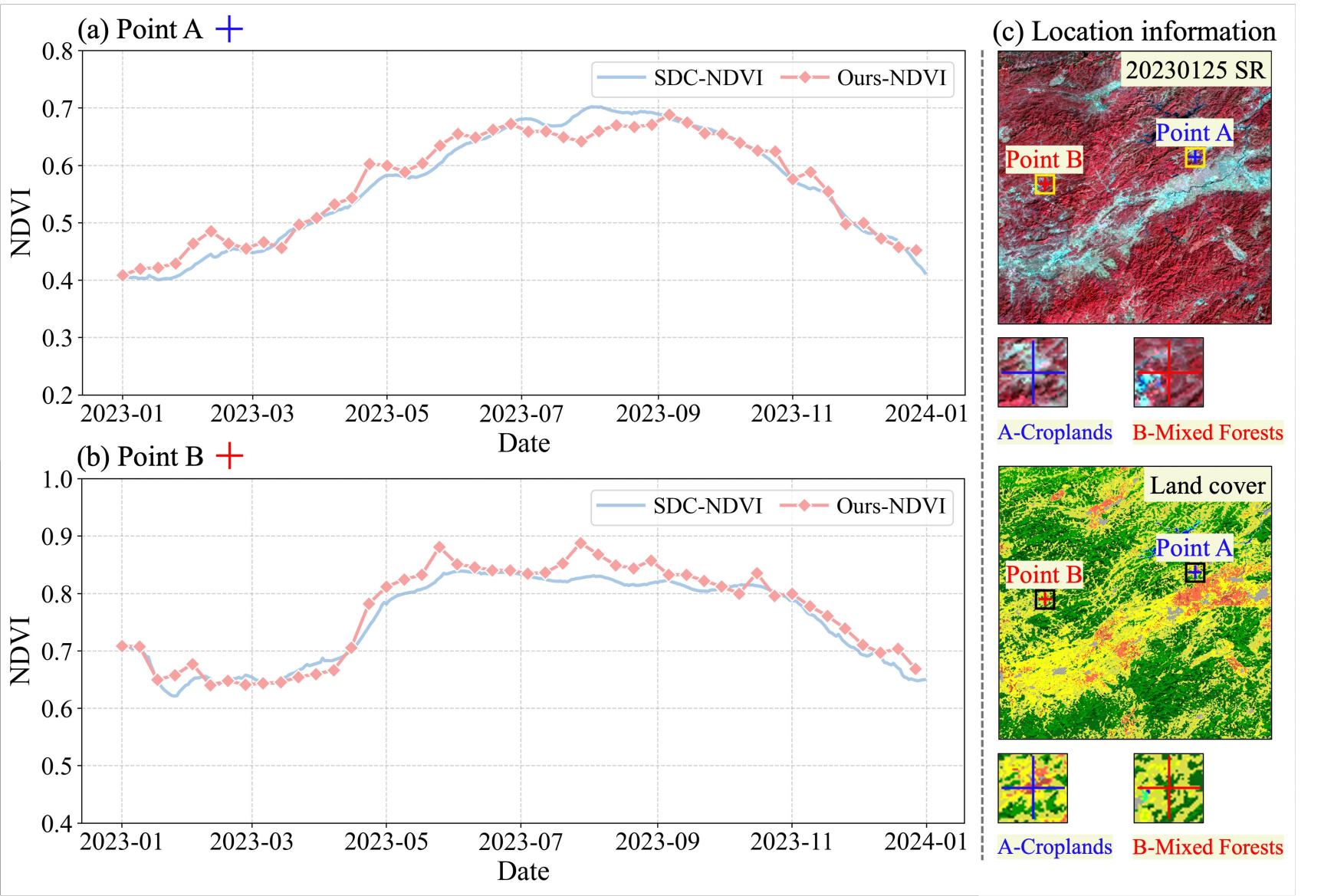}
  \caption{Time series NDVI curves for two sample points and their spatial locations. (a) Point A--croplands. (b) Point B--mixed forests. (c) Spatial locations and land cover information of two points.}
  \label{fig:22NDVI_AB}
\end{figure*}

Time-series-continuous and high-quality SR imagery is essential for vegetation growth monitoring. The normalized difference vegetation index (NDVI) is widely used to characterize vegetation conditions. Therefore, NDVI time series provide an effective way to evaluate both the applicability of reconstructed imagery for vegetation monitoring and the temporal spectral fidelity of reconstructed SR data. To this end, the R13 region in southeastern China is selected as a case study. Located in a subtropical humid monsoon climate zone, this region suffers from severe missing observations throughout the year, providing a challenging scenario for time series reconstruction.

Fig.~\ref{fig:21NDVI} shows the SR images before and after reconstruction on selected dates, together with the corresponding NDVI maps. The NDVI derived from the original observations exhibits pronounced spatial discontinuities and missing values; in missing regions, NDVI is often underestimated and may approach zero. In contrast, the NDVI derived from the reconstructed imagery is spatially and temporally continuous and follows a typical low--high--low seasonal trajectory.

To further examine the temporal spectral fidelity of the reconstructed SR data, the global 0.05$^\circ$ MODIS NDVI seamless data cube (SDC-NDVI) is used as an external reference. This product is derived from the global 500 m MODIS SR seamless data cube \cite{liang2023global}. Two sample points representing croplands and mixed forests are selected for comparison. Fig.~\ref{fig:22NDVI_AB}(a)--(b) presents the daily SDC-NDVI curves and the NDVI curves derived from LC-TSRNet reconstructions for the two sample points. Point A is located in croplands in Quzhou City, Zhejiang Province, China, where double-cropping rice is widely cultivated. Early rice is typically transplanted from mid-to-late April to early May and harvested from late July to early August, followed by the transplantation of late rice. Harvesting generally begins in November. Point B is located in mixed forests and exhibits a smooth seasonal cycle, and NDVI remains relatively high for much of the year. Overall, the reconstructed NDVI agrees well with the SDC-NDVI product in terms of both magnitude and seasonal dynamics, indicating that LC-TSRNet can reliably preserve temporal spectral fidelity and characterize vegetation phenology across different land cover types.

\subsection{Limitations and Future Work}

Although LC-TSRNet performs well in MODIS SR time series reconstruction, it still has several limitations. First, the land cover product and the SR product are mismatched in temporal scale. MCD12Q1 is provided annually, whereas MOD09A1 is an 8-day composite product with 46 observations per year. The ablation results in Section~\ref{Ablation} confirm that a single annual land cover map can provide effective prior information; however, abrupt intra-annual land surface changes, such as seasonal snow/ice cover and wildfire disturbances, cannot be fully captured. This temporal mismatch may introduce biases into the reference SR and subsequently affect reconstruction accuracy.

Second, although the proposed method shows robustness under large-area, persistent, and overlapping gaps, long periods of complete data gaps remain challenging, such as those frequently observed in the Amazon Basin. When valid observations are absent for extended periods, information from neighboring dates and spatial contexts becomes insufficient. Under such conditions, reconstruction relies more heavily on the reference SR derived from the land cover map, which may lead to deviations from the actual surface conditions.

Future work will focus on addressing these limitations. Additional prior information, such as meteorological variables and topographic features, will be incorporated to further improve reconstruction fidelity. In addition, the dataset will be expanded to include a broader range of challenging global scenarios, such as prolonged complete cloud cover and seasonal snow/ice contamination, with the aim of further enhancing the generalization capability and robustness of the proposed method in real-world applications.

\section{Conclusion}
\label{Conclusion}

In this study, a land-cover-guided two-stage method (LC-TSRNet) is proposed for seamless reconstruction of annual MODIS SR products. In the first stage, an annual land cover map is introduced, and the CSPM is designed to transform land cover classes into reference SR time series, thereby providing prior spectral constraints for missing regions and generating coarse representations. This alleviates the lack of reliable prior constraints in MODIS SR reconstruction. In the second stage, the SMRM adaptively exploits and integrates multi-scale spatiotemporal contextual information through parallel multi-scale branches and a progressive aggregation mechanism, enabling the effective reconstruction of large-area, persistent, and overlapping gaps. Subsequently, the 3DAM imposes cross-dimensional joint constraints through three parallel-cascaded attention submodules, thereby enabling the collaborative utilization of spatio-temporal-spectral information in SR time series. As a result, seamless SR time series with rich texture details and high spectral consistency are generated.

In addition, considering the lack of publicly available MODIS sample datasets, the AIRMOD-TSR dataset is constructed. It contains annual MODIS SR time series and corresponding annual land cover maps for 13 representative ROIs worldwide. Extensive experiments on AIRMOD-TSR demonstrate that LC-TSRNet outperforms the comparison methods in reconstruction accuracy and exhibits stronger generalization capability and robustness. The vegetation monitoring application further verifies the temporal spectral fidelity and practical value of the proposed method.

\bibliographystyle{IEEEtran}
\bibliography{reference}

\newpage

 





\end{document}